# DEEP REGRESSION ENSEMBLES

**Antoine Didisheim**
Swiss Finance Institute
University of Lausanne
antoine.didisheim@unil.ch

**Bryan Kelly**
Yale University and
AQR Capital Management
bryan.kelly@yale.edu

**Semyon Malamud**
Swiss Finance Institute
EPFL
semyon.malamud@epfl.ch

March 10, 2022

## ABSTRACT

We introduce a methodology for designing and training deep neural networks (DNN) that we call "Deep Regression Ensembles" (DRE). It bridges the gap between DNN and two-layer neural networks trained with random feature regression. Each layer of DRE has two components, randomly drawn input weights and output weights trained myopically (as if the final output layer) using linear ridge regression. Within a layer, each neuron uses a different subset of inputs and a different ridge penalty, constituting an ensemble of random feature ridge regressions. Our experiments show that a single DRE architecture is at par with or exceeds state-of-the-art DNN in many data sets. Yet, because DRE neural weights are either known in closed-form or randomly drawn, its computational cost is orders of magnitude smaller than DNN.

***Keywords*** Deep learning · Neural network · Random features · Ensembles

## 1 Introduction

Deep neural networks (DNN) have revolutionized machine learning and have impacted many aspects of society. They outperform other learning methods in a vast range of applications, including image recognition, natural language processing, robotics, and games. The standard algorithm for training DNN is stochastic gradient descent (SGD). This requires significant computing power, typically more than available on a desktop computer and often in the form specialized hardware like graphics processors. The properties of DNN trained by SGD are complex, and our understanding of these properties is in its infancy.

Compared to DNN, kernel methods (reproducing kernel Hilbert spaces) are simpler and more transparent, and hence are often used to benchmark DNN performance. Perhaps the most popular and computationally convenient kernel method is random feature regression [1]. This is equivalent to training a wide, two-layer network with randomly drawn input weights and output weights trained by linear ridge regression. The attractiveness of this method is its analytical tractability and low computational cost, which occur because SGD is replaced by linear regression and randomization.

In this paper, we introduce a methodology for designing and training DNN called "deep regression ensembles" (DRE). It combines the expressivity of DNN [e.g. 2] with the analytical tractability of random feature regression. Each layer $m$ in our network has a double ensemble structure, illustrated in Figure 1. To form the first ensemble, inputs from the incoming layer ($X^{(m)}$) are combined via random weights (drawn from a distribution indexed by $k$) and transformed through an activation neuron. Weights are re-drawn $P$ times, as in random feature regression, producing a set of non-linearly transformed regressors ($z_{k,p}, p = 1, ..., P$). These are in turn combined with output weights estimated by ridge regression. The ridge regression is myopic, targeting the final output without regard to subsequent layers in the network (this step bypasses SGD). We run this regression not once but $L$ times using a different choices for the ridge penalty ($\lambda_l$), producing an ensemble of $L$ predictions. To form the second ensemble, collections of $P$ random weights are re-drawn from $K$ different distributions. In our formulation, random features are normally distributed, but the variance of the weights ($\gamma_k$) differs across ensembles. Each of the $K$ sets of random weight draws feeds into a collection of $L$ ridge regressions described above, for a total of $KL$ predictions in the full, double ensemble.

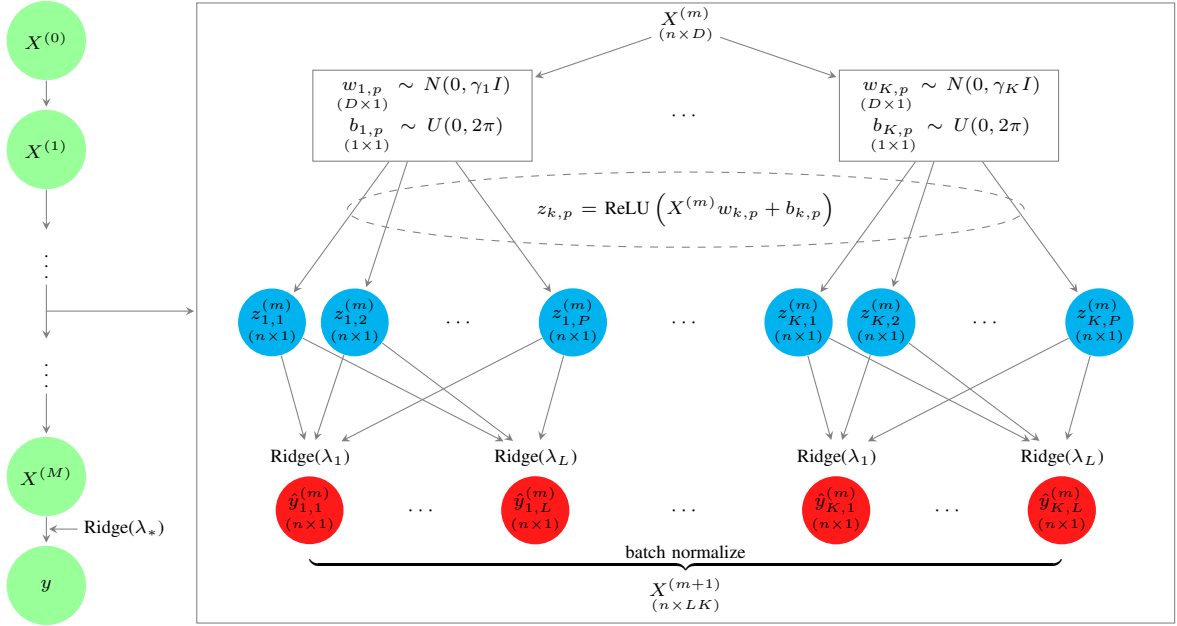

**Figure 1:** The Ensemble Map

Network depth is accommodated by iteratively feeding the ensemble of predictions into a subsequent layer, where they are again transformed into non-linear random features and organized into an ensemble of ridge regression predictions. The left side of Figure 1 shows this process repeated $M$ times, after which one last ridge regression connects the last feature layer ($X^{(M)}$) to the final output, for a network with total depth of $M + 1$.

The two dimensions of the ensemble—one that mixes variances of random features and another that mixes ridge penalties—can be interpreted as two dials that control the model's resolution.[1] The parameter $\gamma_k$ modulates the variability of the input weights and thus the extent of non-linearity extracted by the random features. When $\gamma_k$ is close to zero the features are nearly linear in the original data, while large $\gamma_k$ accentuates the data's non-linear aspects. The ridge parameter $\lambda_l$ modulates variability of the output weights, which can be large and highly expressive (when $\lambda_l$ is near zero) or dampened with heavy shrinkage (when $\lambda_l$ is close to infinity).

DRE is analytically simple and computationally cheap to implement. All neural weights are given by either by a random number generator or in closed form through ridge regression. DRE is trained without SGD and does not use backward feature correction [see, e.g. 3, 4]. Thus, its implementation is orders of magnitudes faster than DNN, does not require specialized hardware, and can be run on a laptop in minutes for moderately sized data sets. In turn, we can train DRE networks of essentially any depth, and can feasibly select among a wide range of DRE architectures (both depth and width) via cross-validation, an infeasible feat for SGD. In particular, DRE enjoys "free tuning" of network depth in the sense that, once a maximum model depth $M$ is estimated (holding width fixed), we can select among all architectures up to depth $M$ with no additional computation.

Our empirical experiments demonstrate that DRE is on par with or exceeds the performance of DNN. We show this first in the context of a classic dataset in which DNN excels, FMNIST [5]. We also demonstrate the beneficial performance of DRE relative to DNN in data simulated from a single-neuron network. This data generating process is an especially interesting proving ground for DRE because theoretical results ensure that, in this environment, DNN is guaranteed to outperform a wide two-layer network estimated with random feature regression or neural tangent kernels [6, 7]. Nonetheless, DRE's predictive performance in these simulations tends to dominate DNN but with computational cost similar to random feature regression. Throughout our empirical analyses we implement a single, universal, and rather simple DRE neural architecture, while the state-of-the-art DNN we compare against require specialized architectures that differ dramatically across datasets. This result, that DRE seems to feature *architecture universality*, is a surprising empirical fact that warrants further theoretical investigation.

---

[1]This is analogous to architectural choices in convolutional nets that visualize image contents with various degrees of resolution.

Lastly, we provide a theoretical motivation for the performance gains from ensembles of random feature ridge regressions that we observe in the data. Our theoretical setting focuses on high complexity ridge models and uses random matrix limit theory. We contrast two ridge regression approaches. One is standard ridge using all $P$ features in a single "flat" regression. The other splits the $P$ features into $K$ subsets, estimates one ridge regression within each subset, then ensembles the fits from the $K$ regressions. We prove the surprising result that, in the high complexity regime, the ensemble approach enjoys large improvements in risk (test mean squared error) versus the single flat ridge regression. Furthermore, DRE solves a well known limitation of random feature regression that its computational complexity is cubic in the number of random features, often causing RAM limits to bind. It is easy for DRE to entertain model specifications unthinkable for random feature regression (without requiring additional memory) because it splits its analysis into many smaller ridge regressions that it subsequently combines in an ensemble.

Interestingly, this is related to a result posed by [8]. They examine ensembles of random neural networks to circumvent the implementation challenges of training extremely large networks. They prove that the loss from such "random kitchen sinks" is small. Building on recent theoretical advances in the analysis of large-dimensional regressions, we prove that this "small loss" *is actually a gain*. Specifically, we show that the risk of the ensemble is always smaller than the risk of the flat model when complexity is sufficiently high. The improvement in risk through ensembling is very large in some conditions, and is monotonically increasing in model complexity. This theoretical result is consistent with experimental evidence for large datasets (see, for example, [9]) suggesting that ensembling leads to significant gains.

As in [8], our theoretical results concern linear ensembling techniques. By contrast, DRE is a form of DNN and is thus a highly non-linear technique through its iterated non-linear composition of prediction ensembles. Furthermore, DRE ensembles not only different subsets of signals, but also ensembles across different model resolutions, which are governed by a distribution of random weight variances and a distribution of ridge penalties. Along these lines, we prove that an ensemble of ridge regressions with different penalties achieves a form of optimal non-linear shrinkage of the entire eigenvalue distribution. This is in contrast with standard flat ridge, which penalizes all eigenvalues of the feature covariance matrix equally (by adding a constant). On the whole, theoretical underpinnings of DRE require further development and this is an important direction for future research. Fortunately, DRE (unlike SGD) enjoys analytical tractability, providing a key opportunity for advancing our understanding of deep learning.

## 2 Related Literature

Our analysis relates to a number of recent and rapidly growing literatures in machine learning. First, the random features model has become a workhorse mathematical framework for understanding generalization and approximation properties of neural networks. We build on the analysis of "double descent" (or "benign overfit") that has recently garnered much attention in the literature.[2] Recent theoretical findings[3] imply that any neural network with infinitely wide layers trained by scaled SGD ("lazy training") can be approximated by a kernel method with an appropriately chosen kernel (the so-called "neural tangent kernel," or NTK).

The hope, based on this literature, is that depth might become unnecessary and shallow but infinitely wide random networks (which are easier to estimate than DNN) can attain the minimal prediction risk. Prompted by this idea, the literature directly compares the performance of easier-to-use kernel methods against DNN, and indeed a number of studies show that DNN can be matched by kernels.[4]

Unfortunately, there is also evidence challenging this view. Kernel methods face a performance ceiling in some environments, and a number of recent papers document substantial outperformance of DNN over the corresponding NTK.[5] Additionally, recent theoretical work characterizes large classes of functions for which kernel methods fail to efficiently approximate the underlying function, leading to underperformance relative to DNN.[6] For example, kernel methods often fail at learning a simple, single-neuron function [6, 7] and hierarchical compositions of simple functions [3, 4]. Relatedly, [41] show that kernel methods are ill-suited for handling isotropic raw features (i.e., when their covariance matrix is close to the identity). In these conditions, kernel methods suffer from a curse of dimensionality, while neural networks are able to accurately approximate the correct low-dimensional representation. [41] demonstrate that as features become noisier (and thus more isotropic), the performance gap between DNN and kernel methods widens. A summary perspective on these findings is that random feature regressions can only recover *projections* of the true function on a subspace (see [42]). However, the subspace is model-specific, so any individual kernel model will be limited by its partial approximation of the true process.

---

[2]See, for example, [10]. Our theoretical results relate to a quickly growing literature on random matrix methods in machine learning, such as [11, 12, 13, 14, 15, 16]. Several of these papers derive explicit expressions for the generalization error of random two-layer networks as a function of the network architecture. We believe that the methods from these papers can be extended to derive analogous results for DRE.

[3]See [17, 18, 19, 20, 21, 22, 23].

[4]See, e.g., [24, 25, 26, 27, 28, 29, 30, 31].

[5]See, e.g., [32, 33]; in particular, a different scaling can lead to significantly improved effectiveness of DNN trained with SGD, as shown by [34, 35, 36, 37].

[6]See, for example, [38, 6, 7, 39, 3, 4, 40].

These are incisive challenges for kernel learning, and DRE offers a rejoinder. We demonstrate that kernel methods, when combined as a deep network of iterated random feature regression ensembles, is an effective learning algorithm even when faced with single-neuron data generating processes and isotropic features. Our experiments show that DRE learns narrow neural representations at least as successfully as DNN. And through experiments that add increasing amounts of noise to our features we demonstrate that DRE overcomes isotropy, systematically matching or outperforming competing models in every experiment we analyze. Most strikingly, this happens with no tuning, with minimal computation time (less than a minute on an 8-core laptop computer), and using the same simple architecture for all data sets. Why does this work? DRE ensembles multiple shallow models, and iterates this process hierarchically. We speculate that, in doing so, it recovers projections of the true function on many different subspaces. If these subspaces are sufficiently different, combining their projections can produce more accurate approximations of the true function.[7] The success of DRE in our numerical experiments supports this intuition.

Finally, our paper contributes to a literature on training neural networks without SGD.[8] In this area, DRE is closest to layer-wise DNN training,[9] as our architecture is also composed of blocks with each block trained locally. However, our approach to constructing and training blocks is fundamentally different from previous work. First, our architecture alternates linear and non-linear sub-layers, and only trains the linear layers (using ridge regression). Second, we use a vast number of output neurons to construct a double ensemble in each layer. Third, the prediction ensemble is passed to subsequent layers where it is again non-linearly transformed. All three aspects of DRE—ensembling over the random weighting model, ensembling over the ridge penalty, and incorporating network depth through composition of ensembles—are crucial for model performance.

## 3 Deep Regression Ensembles Methodology

This section describes how we design and train DRE. We consider a dataset $(y_i, x_i)$, $i = 1, ..., n$ of $n$ observations with labels $y_i \in \mathbb{R}$ and raw (or "intrinsic") features $x_i \in \mathbb{R}^d$. We use $X = (x_i) \in \mathbb{R}^{n \times d}$ to denote the training sample feature matrix, and $y = (y_i) \in \mathbb{R}^n$ to denote the vector of labels in the training sample.

### 3.1 Anatomy of a DRE Layer

We recursively define the DRE model beginning with inputs $X^{(m)}$ inherited as the output from the previous layer (and with $X^{(0)} = X$ initiating the recursion). Let $D$ denote the dimension of the inherited features, so that the input training data matrix $X^{(m)}$ is dimension $n \times D$.

As illustrated in Figure 1, each DRE layer consists of two sub-layers. The first sub-layer uses a non-linear random feature map to convert inputs $X^{(m)}$ into a large number of intermediate features. We generate not one, but $K$ different sets of random features, each having dimension $P$. In particular, random feature $p = 1, ..., P$ in subset $k = 1, ..., K$ is given by[10]

$$z_{k,p}^{(m)} = \text{ReLU}(X^{(m)}\omega_{k,p}D^{-1/2} + b_{k,p}) \quad \text{with} \quad \omega_{k,p} \sim N(0, \gamma_k I), \; b_{k,p} \sim U(-a, a), a > 0\,. \quad (1)$$

In other words, the input weights in each random feature set $k$ are generated by a different Gaussian kernel whose variance is an element of $\Gamma = (\gamma_1, ..., \gamma_K)$. The biases are identically uniformly distributed across all feature sets. Varying $\gamma_k$ allows different feature sets to capture varying degrees of non-linear associations among the input features. Together, $\{z_{k,p}^{(m)}\}$ (the collected blue nodes in Figure 1) constitute an ensemble of random features from heterogeneous kernels.

In the second sub-layer, we combine each set of random features into predictions through a set of myopic ridge regressions. Ridge regressions are run separately in each random feature set $k$, and are run multiple times using a grid of ridge parameters $\Lambda = (\lambda_1, ..., \lambda_L)$. Note that in this structure the network is *not* fully connected. This is crucial for the performance of the network as it preserves heterogeneous behavior across blocks in the ensemble so that they can benefit each other when recombined later in the network. These regressions are myopic in the sense that weights are assigned to random features based on directly predicting the final output label without regard to subsequent layers of the DRE network. Considering a range of ridge parameter allows the model to consider feature weights throughout the spectrum of bias-variance tradeoffs.

---

[7]Other papers find benefits from ensembling, particularly in the presence of large label noise. See, e.g., [43].

[8]A number of papers propose clever algorithms for training DNN without SGD, including [44, 45, 46, 47]. The standard course in these papers is to first select a DNN architecture, then propose an algorithm that is superior to SGD for training the network. These methods, in one way or another, implement a form of backward feature correction.

[9]See, e.g., [48, 49].

[10]Recall that ReLU$(x) = \max(x, 0)$.

Let $Z_k^{(m)} = (\lambda_{k,p}^{(m)})$ denote the $n \times P$ matrix of training data for random feature set $k$. For a ridge parameter $\lambda_l$, the corresponding regression coefficients (and thus estimated network weights) on $Z_k^{(m)}$ are

$$\hat{\beta}_{k,l}^{(m)} = \left(\lambda_l I + \frac{1}{n} Z_k^{(m)\top} Z_k^{(m)}\right)^{-1} \frac{1}{n} Z_k^{(m)\top} y, \qquad (2)$$

and the corresponding prediction is

$$\hat{y}_{k,l}^{(m)} = Z_k^{(m)} \hat{\beta}_{k,l}^{(m)}. \qquad (3)$$

The collection of $KL$ predictions, $\{\hat{y}_{k,l}^{(m)}\}$, constitutes a double ensemble. It gives the model an ability to mix its learning efforts over varying degrees of feature non-linearity ($\gamma_k$) and varying degrees of weight shrinkage ($\lambda_l$). These degrees of freedom give DRE two dials to modulate the resolution of the model. The ensemble collects together a spectrum of models ranging from mostly linear features with heavily constrained weights to rich non-linear specifications with highly expressive weights.

Next, we introduce a form of batch normalization in which each $n \times 1$ component of the prediction ensemble $\{\hat{y}_{k,l}^{(m)}\}$ is normalized by its uncentered standard deviation. We collect the resulting normalized ensemble in the $n \times KL$ matrix $X^{(m+1)}$. If $m$ is an intermediate layer of the network, then $X^{(m+1)}$ is passed as input to the next layer, where the process of double ensembling random feature ridge regressions begins again. If $m$ is the final layer before the output labels, then we run a final ridge regression of $y$ on $X^{(m+1)}$ using a single ridge parameter, $\lambda_*$. The final penalty is the only tuned hyperparameter in our baseline framework, and in our experiments we use a simple test-validation split to select $\lambda_*$.

### 3.2 Comments

Algorithm 1 in the appendix summarizes the design and training of DRE. As the algorithm and Figure 1 illustrate, DRE is an $(M+1)$-layer neural network. But there are several important aspects that differentiate DRE from standard DNN. First, within each layer, input weights are randomly generated and output weights are trained myopically with ridge regression. This makes DRE a direct multi-layer analog of [1]. In addition to the obvious computational benefits, the analytical tractability of DRE makes it amenable to rigorous mathematical analysis using recent theoretical advances in understanding linear ridge regression with random features. Second, due to myopic training, DRE does not use back-propagation which drastically reduces the computational cost. Third, training within a DRE layer is massively parallelizable because each sub-model of the ensemble is trained separately. Fourth, unlike DNN, we produce a usable prediction after each layer. An attractive implication of this is "free tuning" of network depth.

The primary computing cost for DRE is paid to invert high-dimensional covariance matrices for ridge regression, which must be repeated for a grid of $\lambda$-values. We can significantly reduce this cost using eigenvalue decompositions. If $\lambda \in \mathbb{R}^{n \times P}$ and $z^\top z = U \operatorname{diag}(\mu) U^\top$ (with $\mu$ the vector of eigenvalues), then the regression coefficients $\hat{\beta}(\lambda)$ for a grid of $\lambda$ can be computed as

$$(\lambda I + n^{-1} z^\top z)^{-1} n^{-1} z^\top y = U \operatorname{diag}(\mu + \lambda)^{-1} U^\top z^\top y.$$

Therefore, we compute the eigenvalue decomposition once, calculate $\tilde{y} = U^\top z^\top y$ once, then we need only compute matrix multiplication $U \operatorname{diag}(\mu + \lambda)^{-1} \tilde{y}$ for each element of the $\lambda$ grid to complete the necessary inverse calculations. Finally, in a high complexity regime where the number of features is larger than $n$, we can work with $zz^\top$ instead of $z^\top z$ as these matrices have the same non-zero eigenvalues and their eigenvectors are directly linked through multiplication by $z$.

### 3.3 Generalizations

Our data experiments utilize the specific DRE formulation outlined above with fixed choices for network depth $M+1$ and ensemble dimensions $K$ and $L$ that are the same in all layers. We briefly note a few ways that our specific formulation can be generalized at low cost. First, our use of Gaussian weights with a relu activation in equation (1) can easily be replaced by any choice of $K$ heterogenous kernels.[11] Second, it is easy to introduce ensemble the dimensions $K_m$ and $L_m$ to vary the neuron configuration layer-by-layer, $m = 1, \cdots, M$. Lastly, thanks to the computational simplicity of DRE, the dimensions $M$, $K$, and $L$ can be tuned as hyperparameters through cross-validation.

## 4 Experiments

In this section, we report the performance of DRE on simulated and real datasets and compare it with other state-of-the-art models.

### 4.1 Simulated Data

In this set of experiments, we simulate data sets from data generating processes that are provably harder for kernel methods to learn than they are for DNN.

The first simulation experiment is motivated by [6], who show that a random feature regression is ineffective in learning a trivial network with a single ReLU neuron. We thus compare the empirical performance of DRE to DNN and random feature regression in simulated samples from a single-neuron network. Specifically, we consider an activation $\sigma()$ that takes one of two forms, either ReLU or sigmoid. We draw a matrix of *i.i.d.* input data $X \sim N(0, I) \in \mathbb{R}^{n \times d}$. Next, we draw a $d$-vector of weights $w \sim N(0, I)$ truncated at $\pm 3$. The labels are then generated as $y = \sigma(Xw) + \epsilon$ where $\epsilon \sim N(0, s^2 I) \in \mathbb{R}^{n \times 1}$ is *i.i.d.* with standard deviation $s$.

Our experiments fix the simulated sample size at $n = 3000$, feature set size at $d = 50$, and use noise standard deviation $s \in \{0.1, 0.2, ..., 0.9\}$. In each simulated data set and each experiment, we split the sample into three equal size samples for training, validation, and testing. We use validation performance to select the model specification and report out-of-sample performance on the test set.

We implement DRE with the following settings. The number of features in each random feature block of an ensemble is $P = 100$. The number of such blocks is $K = 500$, with each block $k$ having a weight variance $\gamma_k$ sampled from $U(0.25, 1.25)$. Finally, our ensembles use a grid $\Lambda = \{\lambda_i\}_{i=1}^{L}$ of $L = 29$ different ridge penalties.[12]

We compare DRE to three benchmark methods. The first is DNN, which is directly motivated by the structural comparisons between DRE and DNN that we outline above. The second is random feature regression, which is a natural benchmark given its role at the core of DRE and its demonstrated competitiveness with DNN in a number of empirical applications. The third is random forest, motivated its robust success in a wide range of classification problems [50]. For DNN and random forest we choose the best performer in the validation set from several potential architectures. For random feature regression, we set the number of features equal to the size of the final layer in DRE, $K \cdot L = 500 \cdot 29$. Appendix B contains training details for each benchmark algorithm.

Figure 2 reports the risk (which we report as a normalized MSE, i.e., 1-$R^2$) realized by each method. Panel (a) shows results when the single-neuron activation is ReLU. At low levels of noise, DRE matches the performance of DNN, and both of these dominate random feature regression and random forest by a large margin. At higher levels of noise, DNN suffers relative to DRE, with DRE becoming the standalone best performer. Panel (b) shows performance when the data is generated from a sigmoid activation function. Here DNN and DRE are roughly on par when noise is low, but DNN suffers relative to DRE in noisier systems. The pattern for random features regression versus DRE is the opposite. In low-noise data, random features cannot compete with DRE or DNN, but when noise is large, DRE and random feature regressions are roughly on par (and better than DNN).

### 4.2 FMNIST

We next analyze the empirical performance of DRE on real data taken from FMNIST [5]. To adapt our regression framework for categorical data, we conduct ten different experiments using data sets with binary outcomes.

The data for each experiment uses a pair of categories (e.g., "T-shirt/Top" versus "Trouser"). These pair-wise comparisons have 0/1 labels that can be handled with DRE's regression structure, and have equal numbers of observations for each label. To keep the number of experiments to ten, we compare each category to the subsequent category in Table 2 of [5] (for the last category, "Ankle boots," we compare it to the first category, "T-shirt/Top.") We compare DRE performance to DNN and random feature regression, using the same estimation procedure described in Section 4.1 and Appendix B. In addition to studying the raw data (noise level = 0), we also examine the performance effects of adding noise to the data following the formulation of [41].

---

[11] A generic positive definite kernel $K(x_1, x_2)$ admits a representation $K(x_1, x_2) = \int f(x_1; \theta) f(x_2; \theta) p(\theta) d\theta$ for some feature map $f(x; \theta)$. Replacing $z_{k,p}^{(m)}$ in (1) with $z_{k,p}^{(m)} = f(X^{(m)}; \theta_{k,p})$ leads to a DRE with arbitrary kernel-based random features.

[12] The specific grid of ridge parameters is $\Lambda = \{0.0001, 0.001, 0.01, 0.1, 1.0, 5.1, 10.1, 15.1, 20.1, 25.1, 30.1, 35.1, 40.1, 45.1, 50.1, 55.1, 60.1, 65.1, 70.1, 75.1, 80.1, 85.1, 90.1, 95.1, 100.1, 1000.0, 2000.0, 5000.0, 10000.0\}$.

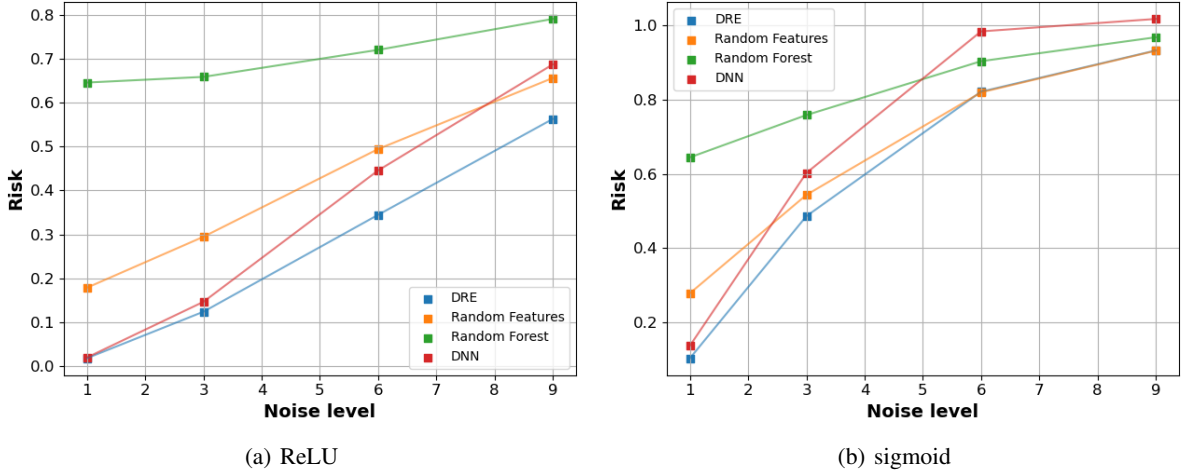

(a) ReLU

(b) sigmoid

**Figure 2:** The figures show out-of-sample performance of DRE and three benchmarks on simulated data. The true function is $\sigma(Xw) + \epsilon$ where $X$ is a $3000 \times 50$ matrix drawn from an *i.i.d.* normal distribution, $w$ is a vector of random weights of dimension 50 drawn from a truncated normal ($|w_i| < 3 \forall i$), $\sigma(\cdot)$ is a ReLU function in panel (a) and a sigmoid function in panel (b), and $\epsilon$ is a vector of noise drawn from a normal with variance between 0.1 (noise level = 1) and 0.9 (noise level = 9).

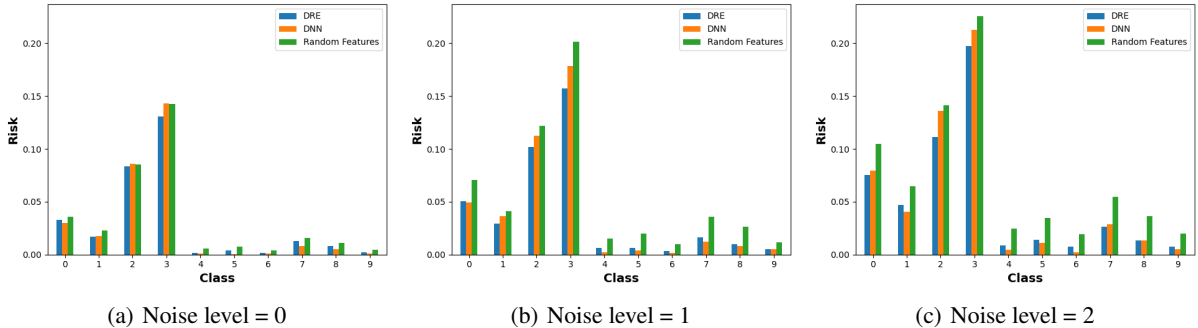

(a) Noise level = 0

(b) Noise level = 1

(c) Noise level = 2

**Figure 3:** The figures show out-of-sample performance of DRE and three benchmarks in FMNIST data with three level of noise. Each set of bars 0-9 represents subset of FMNIST data for a pair of categories (with labels converted to 0/1). We add varying degrees of noise to the data following [41].

The results, shown in Figure 3, suggest three main conclusions. The first is that both DRE and DNN outperform random feature regression in every experiment. The second is that DRE achieves similar performance to DNN across the board. But DRE achieves this performance while reducing computation cost by a few orders of magnitude. Lastly, DRE tends to outperform DNN in the most difficult-to-learn categories, and in these cases the outperformance is comparatively large. On the other hand, when DNN outperforms DRE, the difference is small.

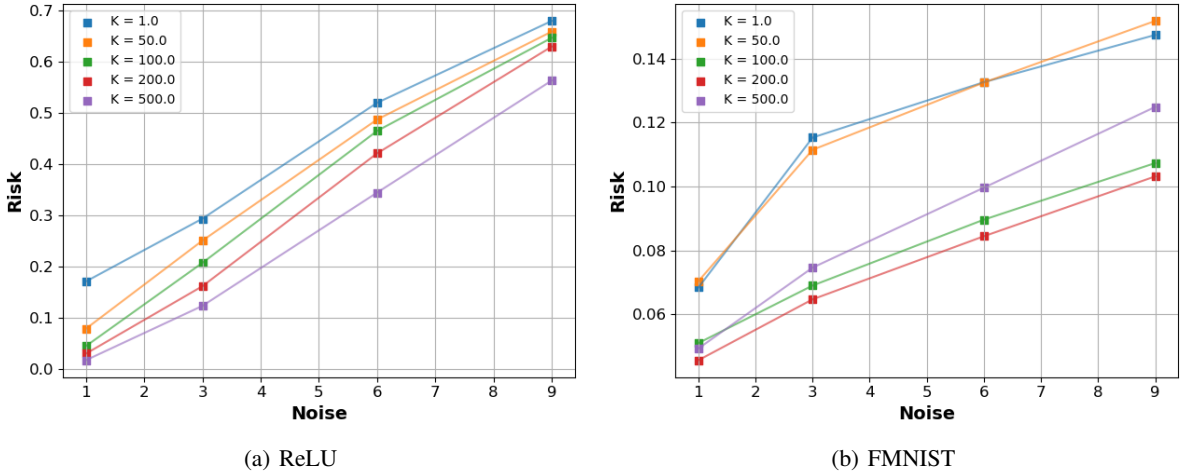

(a) ReLU

(b) FMNIST

**Figure 4:** The figures illustrate the role of ensembling and report DRE's performance sensitivity to the choice of $K$. Panel (a) shows results for single-neuron ReLU simulations from Figure 2 and Panel (b) shows FMNIST data from Figure 3.

### 4.3 Distinguishing the Benefits of Ensembling and Depth

DRE has two primary architectural components: random feature ensembles and iterated non-linear compositions (i.e., depth). First, we modify the simulation analysis reported in Figure 2 in order to isolate the contribution of ensembling to DRE's performance.

In the simulation analysis of Figure 2, the total number of random features in each DRE layer is fixed at $P \cdot K = 50000$, and in the FMNIST experiments of Figure 3 use $P \cdot K = 300000$. To understand the role of ensembling, we hold the total feature count in these experiments fixed at their previous values, but divide it into $K = 1, 50, 100, 200, 500$ sub-models and re-run the DRE algorithm. We also consider how the role of ensembling is affected by varying the noise environment (as in our preceding experiments).

Figure 4 reports the results for the single-neuron ReLU simulation in Panel (a) and FMNIST[13] data in Panel (b). DRE performance tends to improve by dividing features into a large number of sub-models, though the effect can be non-monotonic. The benefits of ensembling are large at all noise levels, and corroborate the theoretical predictions of Section 5. It is also worth noting that when we divide a fixed number of random features into more and more sub-models, the total computational cost of DRE drops, providing an additional benefit of ensembling beyond the gains in prediction accuracy.

Next, we investigate how the performance of DRE varies with network depth. Figure 5 reports DRE specifications with depth $M = 1, ..., 5$, holding other network attributes fixed. Performance for single-neuron ReLU simulations in Panel (a) and FMNIST data in Panel (b) improves when moving beyond a $M = 1$. The effects are larger in low-noise data. They are also non-monotonic, as the additional parameterization due to more layers can eventually undermine out-of-sample performance, as seen in Panel (a).

## 5 Theoretical Results: Linear Ensembling

We consider a simple setting with $p$ features split into $K$ groups, and features have limited dependence across groups. We are interested in estimator behavior in the high complexity regime where $p$ is large relative to the number of observations $n$ (fixed $K$, $p, n \to \infty$, and $p/n \to c > 0$). We prove that, in this environment, a single large ridge regression using all features is inefficient. The regression involves inverting a giant covariance matrix of all features, and the cross-group covariance terms are dominated by noise. In the limit of high model complexity, the effect of this noise can severely inhibit the performance of the large joint ridge regression. We contrast this with ensembling; in particular, we consider the alternative of estimating separate ridge regressions for each feature group and then combining forecasts. Ensembling has provably better performance, and in some circumstances the improvements from ensembling are drastic. That $K$ is fixed means that ensemble gains accrue when the ridge sub-models in the ensemble are themselves complex (in other words, the ensemble can underperform when features are subsetted into very fine groups). The gap between the flat ridge regression and the ensemble can be computed explicitly using random matrix theory methods from [13] and [51]. This result provides a first motivation for the ensemble structure of DRE.

---

[13]We continue to use the binary FMNIST formulation described earlier and, for the purposes of Figures 4 and 5, use just the dataset corresponding to the first classes (i.e., corresponding to the set of bars labeled 0 in Figure 3).

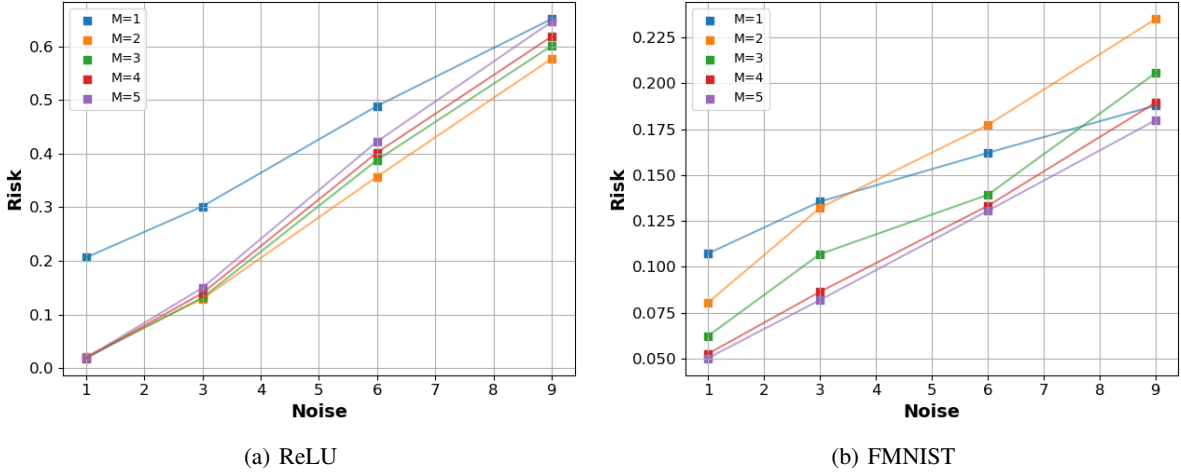

(a) ReLU  (b) FMNIST

**Figure 5:** The figures illustrate the role of depth and DRE's performance sensitivity to the choice of $M$, fixing $K = 500$ and $P = 100$, and at varying levels of noise. Panel (a) shows results for single-neuron ReLU simulations from Figure 2 and Panel (b) shows FMNIST data from Figure 3.

Our second theoretical observation is that selecting a single ridge penalty can also be inefficient. Ensembling ridge regressions that use different ridge penalties can lead to further reductions in model risk, and motivates the multiple-penalty aspect of our DRE architecture. Here again we use random matrix theory methods from [13] and [51] to explicitly characterize the corresponding gains in predictive ability.

### 5.1 Data and Risk

Building on the baseline notation in Section 3, we consider i.i.d. data $(y_i, x_i)$, $i = 1, ..., n$, with $x_i \in \mathbb{R}^p$ being the feature vector and $y_i \in \mathbb{R}$ the label. We use $X = (x_i) \in \mathbb{R}^{n \times p}$ and $y = (y_i) \in \mathbb{R}^n$ to denote the stacked features and labels in the $n$ training observations. The large, flat ridge regression (the "flat model" for short) uses the flat feature vector $x_i$.

We use the notation $x_i(k) \in \mathbb{R}^{p(k)}$ to denote the subset of $p(k)$ features that flow into the $k^{th}$ component of the regression ensemble, which has training matrix analogue $X(k) \in \mathbb{R}^{n \times p(k)}$. The sets $x_i(k)$, $k = 1, ..., K$, are disjoint and exhaust the elements $x_i$.

We assume that

$$y_i = x_i^\top \beta + \varepsilon_i = \sum_{k=1}^{K} x_i(k)^\top \beta_i(k) + \varepsilon_i, \ i = 1, ..., n, \tag{4}$$

where $\beta$ is the coefficient on the flat feature vector and $\beta(k)$ is the subset of coefficients corresponding to $x_i(k)$. The flat feature vector $x_i$ is i.i.d. with $E[x_i] = 0$, $\text{Cov}(x_i) = \Sigma$, with covariance sub-matrices $\text{Cov}(x_i(k_1), x_i(k_1)) = \Sigma(k_1, k_2)$. The errors $\varepsilon_i$ are independent of $x_i$, $E[\varepsilon_i] = 0$, and $\text{Var}[\varepsilon_i] = \sigma^2$. Without loss of generality, we normalize $\sigma^2 = 1$.

Given an estimator $\hat{\beta} \in \mathbb{R}^p$ and a test sample observation $x_0$ sampled from the same distribution as $x_i$, we investigate the generalization properties of our estimator, described by the out-of-sample prediction risk:

$$Risk(\hat{\beta}; \beta) = E[(x_0^\top \hat{\beta} - x_0^\top \beta)^2 | X] = E[\|\hat{\beta} - \beta\|_\Sigma^2 | X], \tag{5}$$

where we follow the notation of [13] and denote $\|x\|_\Sigma^2 = x^\top \Sigma x$.

## 5.2 Ridge Regression Ensembles

We investigate properties of ridge regression sub-model estimators in feature group $k$

$$\hat{\beta}(\lambda; k) = \left(\lambda I + \frac{1}{n} X(k)^\top X(k)\right)^{-1} \frac{1}{n} X(k)^\top y \qquad (6)$$

as well as "flat models" estimated from all features

$$\hat{\beta}(\lambda; flat) = \left(\lambda I + \frac{1}{n} X^\top X\right)^{-1} \frac{1}{n} X^\top y. \qquad (7)$$

Note that each $\hat{\beta}(\lambda; k)$ is also an estimator of $\beta$ where we shrink all components $\beta(l)$, $l \neq k$, to zero. From now on, without additional explanations, we use the notation $\hat{\beta}(\lambda; k)$ to denote both the vector in $\mathbb{R}^{p(k)}$ and its natural embedding into $\mathbb{R}^p$.

The risk of an ensemble of sub-models with coefficients $\alpha = (\alpha_k)$ and ridge parameters $\lambda_k$, $k = 1, ..., K$ can be decomposed as

$$\begin{aligned}
Risk\left(\sum_k \alpha_k \hat{\beta}(\lambda_k; k); \beta\right) &= E\left[\left\|\sum_k \alpha_k \hat{\beta}(\lambda_k; k) - \beta\right\|_\Sigma^2 \bigg| X\right] \\
&= \sum_{k=1}^K \alpha_k^2 E[\|\hat{\beta}(k) - \beta(k)\|_{\Sigma(k)}^2 | X] \\
&\quad + 2\alpha_{k_1} \alpha_{k_2} \sum_{k_1 < k_2} E[(\hat{\beta}(k_1) - \beta(k_1))^\top \Sigma(k_1, k_2)(\hat{\beta}(k_2) - \beta(k_2)) | X].
\end{aligned} \qquad (8)$$

In general, the asymptotic behavior of off-diagonal terms in (8) can depend in a complex way on the covariance structure of signals [see 51]. When the cross-block covariances $\Sigma(k_1, k_2)$ are not too large, the corresponding elements of the giant sample covariance matrix $X^\top X$ are dominated by noise and, as we show below, this leads to large increases in risk when model complexity is high. To compute the asymptotic gain from ensembling in the large-$n$ limit, we need the following additional technical assumptions.

**Assumption 1** *We have*

$$x = \Sigma^{1/2} z,$$

*where $z \in \mathbb{R}^p$ has zero first and third-order moments, and $E[zz'] = I$. Furthermore, the fourth moments $E[z^4]$ are uniformly bounded and $z$ satisfies the Lindenberg condition*

$$\lim_{p \to \infty} \frac{1}{p} \sum_{i=1}^p E[z_i^2 I_{|z_i| > \varepsilon \sqrt{p}}] = 0 \; for \; all \; \varepsilon > 0.$$

**Assumption 2** *The matrix $\Sigma$ has a block structure, $\Sigma = (\Sigma(k_1, k_2))_{k_1, k_2 = 1}^K$ with $\Sigma(k) \equiv \Sigma(k, k) \in \mathbb{R}^{p(k) \times p(k)}$. For each $k_1 \neq k_2$, we have $\mathrm{tr}(\Sigma(k_1, k_2)\Sigma(k_2, k_1)) = o(p)$ as $p \to \infty$.[14] We use $\mu_i(\Sigma)$, $i = 1, \cdots, p$, to denote the eigenvalues of an arbitrary matrix $\Sigma$. In the limit as $p(k) \to \infty$, the spectral distribution $F^{\Sigma(k)}$ of the eigenvalues of $\Sigma(k)$,*

$$F^{\Sigma(k)}(x) = \frac{1}{p(k)} \sum_{i=1}^{p(k)} \mathbf{1}_{\mu_i(\Sigma(k)) \leq x} \qquad (9)$$

*converges to a non-random probability distribution $H(k)$ supported on $(0, +\infty)$.[15] Furthermore, $\Sigma(k)$ is uniformly bounded as $p(k) \to \infty$. We will use*

$$\sigma_{*,i}(k) = \lim_{p(k) \to \infty} p(k)^{-1} \mathrm{tr}(\Sigma(k)^i), \; i \geq 1$$

*to denote asymptotic moments of the eigenvalues of $\Sigma(k)$.*

Our last assumption governs the behavior of the true predictive coefficients, $\beta$. We assume that the distribution of $\beta$ is isotropic.[16]

---

[14] For example, this is the case when $\Sigma(k_1, k_2)$ have uniformly bounded ranks, in which case $\mathrm{tr}(\Sigma(k_1, k_2)\Sigma(k_2, k_1))$ is also bounded when $\Sigma$ itself is bounded. This assumption guarantees that cross-covariances, $\Sigma(k_1, k_2)$, have no impact on asymptotic risk.

[15] If 0 is in the support of $H$, then $\Sigma(k)$ is strictly degenerate, meaning that some signals are redundant.

[16] As [13] show, in general the asymptotic behaviour depends on the distribution of projections of $\beta$ on the eigenvectors of $\Sigma$.

**Assumption 3** *We assume $\beta(k)$ is random, $\beta(k) = (\beta_i(k))_{i=1}^{p(k)} \in \mathbb{R}^{p(k)}$, independent of $z$ and $\varepsilon$, and satisfies $E[\beta] = 0$, and $E[\beta(k)\beta(k)^\top] = p(k)^{-1}b_{*,p(k)}(k)I$ for some constant $b_{*,p(k)} = E[\|\beta(k)\|^2]$,[17] and satisfies $b_{*,p(k)} \to b_*(k)$ almost surely, for some $b_*(k) > 0$. Furthermore, $E[\beta_i^4] \leq q \cdot p(k)^{-2}$ for some constant $q > 0$, and $\beta$ satisfies the same Lindenberg condition as $z$.*

Randomness of $\beta$ in Assumption 3 is a technical condition that simplifies the analysis. As [13] show, it is possible to derive explicit expressions for asymptotic risk without this assumption, but the expressions become more complex. In this case, the asymptotic behavior depends on the distribution of projections of $\beta$ on the eigenvectors of $\Sigma$ (the feature principal components).[18] We leave this important generalization for future research.

### 5.3 Main Results

In this subsection, we fix the ensemble size $K \geq 1$ and consider the limit as $p(k) \to \infty$ for all $k$ and $n \to \infty$ while $p(k)/n \to c(k)$. We use the following auxiliary objects:

$$\begin{aligned} m_k(-\lambda; c(k)) &= \lim_{p(k),n\to\infty} p(k)^{-1} \operatorname{tr}(\lambda I + \hat{\Sigma}(k))^{-1}, \; \hat{\Sigma}(k) = \frac{1}{n} X(k)^\top X(k) \\ m(-\lambda) &= \lim_{p(k),n\to\infty} P^{-1} \operatorname{tr}(\lambda I + \hat{\Sigma})^{-1}, \; \hat{\Sigma} = \frac{1}{n} X^\top X \end{aligned} \quad (10)$$

These limits exist almost surely [see, e.g. 13, 51]. Furthermore, due to the $\operatorname{tr}(\Sigma(k_1, k_2)\Sigma(k_2, k_1)) = o(p)$ condition in Assumption 2, asymptotic risk coincides with that of a block-diagonal $\Sigma$:

$$m(-\lambda) = \sum_{k=1}^K \bar{c}^{-1} c(k) m_k(-\lambda; c(k)), \quad (11)$$

where $\bar{c} = \sum_{k=1}^K c_k$. Using classic results [see, e.g. 13], we have existence of the following probability limits:

$$\begin{aligned} \xi_k(\lambda; c(k)) &= \lim_{p(k),n\to\infty} p(k)^{-1} \operatorname{tr}(\Sigma(k)(\lambda I + \hat{\Sigma}(k))^{-1}) = \frac{1 - \lambda m_k(-\lambda; c(k))}{c(k)^{-1} - 1 + \lambda m_k(-\lambda; c(k))} \\ \xi(\lambda) &= \lim_{p(k),n\to\infty} P^{-1} \operatorname{tr}(\Sigma(\lambda I + \hat{\Sigma})^{-1}) = \frac{1 - \lambda m(-\lambda)}{\bar{c}^{-1} - 1 + \lambda m(-\lambda)}. \end{aligned} \quad (12)$$

Then, we define

$$\begin{aligned} \nu_k(\lambda; c(k)) &= \sigma_{*,1}(k) - c(k)^{-1}\lambda\xi_k(\lambda; c(k)) &&= \lim p(k)^{-1} \operatorname{tr}(\hat{\Sigma}(k)(\lambda + \hat{\Sigma}(k))^{-1}\Sigma(k)) && > 0 \\ \nu'_k(\lambda; c(k)) &= -c(k)^{-1}(\xi_k(\lambda; c(k)) + \lambda\xi'_k(\lambda; c(k))) &&= -\lim p(k)^{-1} \operatorname{tr}(\hat{\Sigma}(k)(\lambda I + \hat{\Sigma})^{-2}\Sigma) && < 0 \quad (13) \\ \hat{\nu}_k(\lambda; c(k)) &= \nu(\lambda; c(k)) + \lambda\nu'(\lambda; c(k)) &&= \lim p(k)^{-1} \operatorname{tr}(\hat{\Sigma}(k)^2(\lambda I + \hat{\Sigma}(k))^{-2}\Sigma) && > 0. \end{aligned}$$

From here, we compute the asymptotic risk of ridge estimators and their ensembles.

#### 5.3.1 Ensembling With Common Ridge Penalty

We start by characterizing the asymptotic risk (8) of an arbitrary ridge ensemble.

**Proposition 1 (Risk of the Ensemble)** *In the limit as $p(k)/n \to c(k) > 0$, $n \to \infty$, the risk of ensemble in (8) converges to*

$$Risk^*\left(\sum_k \alpha_k \hat{\beta}(\lambda_k; k); \beta\right) = \sum_{k=1}^K Risk^*(\alpha_k \hat{\beta}(\lambda_k; k); \beta(k)), \quad (14)$$

*where*

$$\begin{aligned} Risk^*(\alpha_k\hat{\beta}(\lambda; k); \beta(k)) &= \lim_{\substack{p(k),n\to\infty \\ p(k)/n\to c(k)}} E[\|\alpha_k\hat{\beta}(\lambda, k) - \beta(k)\|^2_{\Sigma(k)}|X] \\ &= b_*(k)\sigma_{*,1}(k) - 2\alpha_k b_*(k)\nu_k(\lambda) \\ &\quad + \alpha_k^2(b_*(k)\hat{\nu}_k(\lambda) - c(k)\nu'_k(\lambda)(1 + \bar{b}_* - b_*(k)\sigma_{*,1}(k))). \end{aligned} \quad (15)$$

---

[17] This identity follows because $b_*(k) = \operatorname{tr} E[\beta(k)\beta(k)^\top] = E[\operatorname{tr}(\beta(k)\beta(k)^\top)] = E[b_*]$.

[18] See, [13]. In particular, when $\beta$ is concentrated on the top principal components, the phenomenon of benign overfit emerges ([52], [53]), and the optimal ridge penalty is zero.

When $\alpha_k = 1$, *risk is minimized by*[19]

$$\lambda_{*k} = c(k)(1 + \bar{b}_* - b_*(k)\sigma_{*,1}(k))/b_*(k)$$

where $\bar{b}_* = \sum_{k=1}^{K} b_*(k)\sigma_{*,1}(k)$.

*For a fixed $\lambda$, risk is minimized by*

$$\alpha_{*k} = \frac{b_*(k)\nu_k(\lambda)}{b_*(k)\hat{\nu}_k(\lambda) - c(k)\nu'_k(\lambda)(1 + \bar{b}_* - b_*(k)\sigma_{*,1}(k))}.$$

*The joint minimum of risk over $(\alpha_k, \lambda_k)$ is attained by $(1, \lambda_{*k})$.*

Proposition 1 has a close connection to DRE. It states that optimal ensemble weights ($\alpha_k$) are equal to one when the optimal ridge penalty for each ensemble is known. Naturally, this is infeasible and the ridge penalty must be inferred from data using cross-validation or restrictions imposed by random matrix theory. However, Proposition 1 also shows that when $\lambda_k$ is fixed at an arbitrary (sub-optimal) value, then optimizing the ensemble weights salvages some of the performance that would be otherwise lost due to imperfect $\lambda_k$. In each DRE layer, we first estimate a range of models using generic (non-optimized) ridge penalties. Then, we combine these models with a second-tier ridge regression that seeks to optimally weight sub-models in the spirit of $\alpha_{*k}$.

We now proceed to computing the risk of the flat model with and without optimal penalty. This will serve as a benchmark against which we compare the ensemble. We focus here on a simplified setting where $p(k) = p$ and $\Sigma(k)$ has an eigenvalue distribution that is independent of $k$ (the general case is somewhat cumbersome and requires solving a fixed point system). For simplicity, we normalize $\sigma_*(k) = 1$ for all $k$. The following is true.

**Proposition 2 (Risk of the Flat Model)** *Suppose that $p(k) = p$ for all $k$ and $p/n \to c$, and that the eigenvalue distribution of $\Sigma(k)$ is independent of $k$. Then, the risk of the flat estimator $\hat{\beta}(\lambda)$ is*

$$Risk^*(\hat{\beta}(\lambda); \beta) = \lim_{\substack{p,n\to\infty \\ p/n\to c}} E[\|a\hat{\beta}(\lambda) - \beta\|_\Sigma^2 | X] = \bar{b}_* - 2a\bar{b}_*\nu(\lambda) + a^2(\bar{b}_*\hat{\nu}(\lambda) - cK\nu'(\lambda)), \qquad (16)$$

*where $a$ is the scale of the prediction. When $a = 1$, the risk is minimized by $\bar{\lambda}_* = cK/\bar{b}_*$. For a fixed $\lambda$, the risk is minimized by*

$$\bar{a}_* = \frac{\bar{b}_*\nu(\lambda)}{\bar{b}_*\hat{\nu}(\lambda) - cK\nu'(\lambda)}. \qquad (17)$$

*The joint minimum of risk over $(a, \lambda)$ is attained by $(1, \bar{\lambda}_*)$.*

### 5.3.2 Numerical Illustration

To provide a direct comparison between the ensemble and flat model, and to provide a numerical illustration, we set $\Sigma(k) = I$ for all $k$.[20] In this case, if the ridge penalty is set optimally for each sub-model to $\lambda_{*k}$[21] (which also implies optimal ensemble weights equal one), then the optimizer of (14) is given by $\sum_k \hat{\beta}(\lambda_{*k}, k)$ and the asymptotic risk of the ensemble is

$$Risk^*\left(\sum_k \hat{\beta}(\lambda_{*k}; k); \beta\right) = \bar{b}_* - \sum_{k=1}^{K} b_*(k)\nu(\lambda_{*k}; c). \qquad (18)$$

Likewise, the risk of the flat model with its own optimal penalty $\bar{\lambda}_*$ is

$$Risk^*(\hat{\beta}(\lambda_*); \beta) = \bar{b}_*(1 - \nu(\bar{\lambda}_*)). \qquad (19)$$

---

[19]The optimality of this $\lambda_{*k}$ follows by the same arguments as in [51].
[20]$\Sigma(k) = I$ leads to the Marcenko-Pastur law:

$$\xi_k(\lambda) = cm(-\lambda; c), \ \xi(\lambda) = cKm(-\lambda; cK),$$

with

$$m(-\lambda; c) = \frac{2}{\sqrt{((1-c)+z)^2 + 4cz} + ((1-c)+z)}, \quad \text{and} \quad m'(-\lambda; c) = \frac{cm^2(-\lambda; c) + m(-\lambda; c)}{2czm(-\lambda; c) + (1-c+z)}.$$

[21]In reality, this would requires estimating $b_*(k)$. It is possible to use random matrix theory to estimate $b_*$ using the training data. Alternatively, one can estimate $\lambda_{*k}$ via cross-validation.

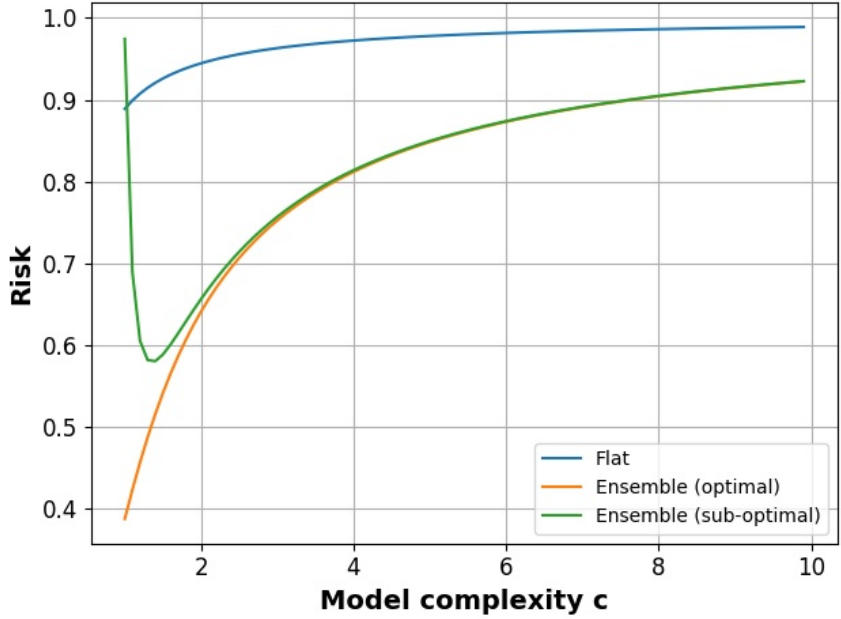

**Figure 6:** The Figure shows theoretical risk of the ensemble and the flat model as a function of model complexity $c$ in the setting of Proposition 3 with $K = 10$. "Flat" is the risk of the flat model (19), "Ensemble (optimal)" is the risk of the optimal combination of sub-models (18), and "Ensemble (sub-optimal)" is risk of the ensemble with a homogeneous penalty of $\lambda_*$.

A interesting question regarding the ensemble is what happens when we run the flat model with its optimal $\lambda_*$, and use *the same* $\lambda_*$ for all sub-models in the ensemble. To answer this, note that for generic (potentially suboptimal) $\lambda$, the optimal ensemble risk with weights $\alpha_{*k} = \alpha_{*k}(\lambda)$ is given by

$$Risk^*\left(\sum_k \alpha_{*k}\hat{\beta}(\lambda;k);\beta\right) = \bar{b}_* - \sum_k \frac{(b_*(k)\nu(\lambda;c))^2}{b_*\hat{\nu}(\lambda;c) - c\nu'(\lambda;c)(1+\bar{b}_* - b_*(k))}. \quad (20)$$

As we now show, this ensemble achieves lower risk than the flat model in the high complexity regime.

**Proposition 3** *Suppose we have heterogeneous $b_*$, so that $\max_k b_*(k) > \min_k b_*(k)$. Then,*

$$Risk^*(\hat{\beta}(\lambda);\beta) > Risk^*\left(\sum_k \alpha_{*k}\hat{\beta}(\lambda;k);\beta\right).$$

*for all sufficiently large $\lambda$.*

Figure 6 illustrates the result of Proposition 3. When complexity is sufficiently high, the ensemble achieves a lower risk than the flat model even with a suboptimal penalty. For low complexity levels, sub-optimal shrinkage can be detrimental and flat model may dominate. Interestingly, at high levels of complexity, the sub-optimal ensemble scheme performs at par with the infeasible optimal ensemble.

Note that the risk of the ensemble and flat model are increasing at sufficiently high levels of complexity. This may at first glance appear to violate the now well established "double descent" property of high-dimensional regression. However, this pattern arises because each value of $c$ on the plot also corresponds to a different choice of optimal ridge penalty, $\lambda_* = c/b_*$, and thus merely reflects that (for correctly specified models) it is harder to learn a more complex model. See, for example, Figure 1 of [13].

### 5.3.3 Non-linear Shrinkage Interpretation of Ensembles With Heterogenous Ridge Penalty

Our final comment on the theoretical motivation of DRE architecture regards its use of heterogeneous ridge penalties. Standard ridge linearly shrinks eigenvalues of the feature covariance matrix through an additive adjustment equal to $\lambda I$.

Consider an ensemble of ridge models with a range of penalties, and define this estimator as

$$\hat{\beta}_{(\Lambda;w)} = \sum_{\ell=1}^{L} w_\ell \hat{\beta}(\lambda_\ell) \qquad (21)$$

for a grid $\Lambda = \{\lambda_\ell\}_{\ell=1}^{L}$ of ridge penalties. Proposition 4 solves the problem of minimizing the risk of this estimator. Interestingly, standard random matrix results imply that the ensemble weights $w$ can be chosen deterministically, independent of $X$.

**Proposition 4** *The solution to the problem*

$$Risk^*(\hat{\beta}_{(\Lambda;w)}; \beta) = \min_{w} \lim_{\substack{p,n\to\infty \\ p/n\to c}} E[\|\hat{\beta}_{(\Lambda;w)} - \beta\|_\Sigma^2 | X] \qquad (22)$$

*is given by*

$$w = \Gamma^{-1}\gamma$$

*where*

$$\gamma = (\nu(\lambda_\ell))_{\ell=1}^{L}, \quad \Gamma = (\Gamma(\lambda_{\ell_1}, \lambda_{\ell_2}))_{\ell_1,\ell_2=1}^{L}$$

*with*[22]

$$\Gamma(\lambda_1, \lambda_2) = b_*\Big(\Sigma_{*,1} + c^{-1}(\lambda_2 - \lambda_1)^{-1}(\lambda_1^2 \xi(\lambda_1) - \lambda_2^2 \xi(\lambda_2))\Big) + (\lambda_2 - \lambda_1)^{-1}(\lambda_2 \xi(\lambda_2) - \lambda_1 \xi(\lambda_1)),$$

*and $\xi$ given by* (13). *The optimal asymptotic risk* (22) *is then*

$$b_* \sigma_{*,1} - \gamma^\top \Gamma^{-1} \gamma.$$

The ensemble (21) admits another interpretation—it achieves *non-linear* eigenvalue shrinkage. To see this, consider a function $f$ and define[23]

$$\hat{\beta}_{f(\cdot)} = f(\hat{\Sigma}) \frac{1}{n} X^\top y, \qquad (23)$$

where ridge regularization corresponds to the choice of $f_\lambda(x) = (\lambda + x)^{-1}$:

$$\hat{\beta}_{f_\lambda(\cdot)} = \hat{\beta}_\lambda. \qquad (24)$$

Now, find the function $f(\cdot)$ attaining minimal risk for the prediction problem. Solving this optimization problem directly is non-trivial, yet a simple ensemble approach can be used to find an approximate solution. Namely, for any vector $w = (w_\ell)_{\ell=1}^{L}$ of coefficients and any grid $\Lambda = (\lambda_\ell)_{\ell=1}^{L}$, we can define

$$f(x; \Lambda; w) = \sum_\ell w_\ell (x + \lambda_\ell)^{-1}. \qquad (25)$$

Any smooth function on a bounded interval can be uniformly approximated by a function $f(x; \Lambda; w)$ with some $\Lambda$ and $w$. Proposition 4 solves the risk minimization problem over all functions in the class (25).

That an ensemble of heterogeneous ridge penalties achieves non-linear shrinkage is a potentially powerful tool in real world applications, and motivates its role in the DRE architecture. Recent advances [e.g. 54, 55] demonstrate that non-linear eigenvalue shrinkage is superior to linear shrinkage when eigenvalues are dispersed (a typical condition observed in empirical covariance matrices).

## 6 Conclusions

We introduce deep regression ensembles (DRE), a novel deep neural architecture that combines the rich expressivity of DNN with the analytical tractability and computational simplicity of random features. In prediction tasks where random features (and kernel methods more generally) are well known to underperform DNN, we show that DRE performs at least as well as DNN while reducing computation cost by orders of magnitude relative to DNN. In addition to being a useful alternative to classic DNN trained by SGD, DRE offer a new road towards a theoretical understanding of deep learning and optimal DNN design based on recent theoretical advances in random matrix theory [11, 12, 13, 14, 15, 16]. Our theoretical results for linear ensembles provide a first step in this direction.

---

[22]Note that the diagonal of this matrix has to be computed as a limit: $\Gamma(\lambda_1, \lambda_1) = \lim_{\lambda_2 \to \lambda_1} \Gamma(\lambda_1, \lambda_2)$ and can be expressed in terms of the derivatives of the functions involved.

[23]As usual, $f(\hat{\Sigma})$ is defined through functional calculus: If $\hat{\Sigma} = U \operatorname{diag}(\mu) U^\top$ is the eigenvalue decomposition of $\hat{\Sigma}$, then $f(\hat{\Sigma}) = U \operatorname{diag}(f(\mu)) U^\top$.

## Acknowledgments

AQR Capital Management is a global investment management firm, which may or may not apply similar investment techniques or methods of analysis as described herein. The views expressed here are those of the authors and not necessarily those of AQR. Semyon Malamud gratefully acknowledges support from the Swiss Finance Institute and the Swiss National Science Foundation.

## A  Proofs

**Proof of Proposition 1.** Throughout the proofs, we use $X \approx Y$ to denote the fact that $X - Y$ converges to zero in probability.

Using that $\beta^\top \Sigma \beta \approx \sum_k \sigma_*(k) b_*(k) = \bar{b}_*$, we can decompose the risk of any estimator $\hat{\beta}$ as

$$E[\|\beta - \hat{\beta}\|_\Sigma^2] \approx \bar{b}_* - 2\beta^\top \Sigma \hat{\beta} + \hat{\beta}^\top \Sigma \hat{\beta} \tag{26}$$

We will use the identity

$$\hat{\beta}(k) = (\lambda I + \hat{\Sigma}(k))^{-1} \frac{1}{n} \sum_i x_i(k)(x_i^\top \beta + \varepsilon_i) = (\lambda I + \hat{\Sigma}(k))^{-1}(\hat{\Sigma}(k,:)\beta + q(k)), \tag{27}$$

where we have defined

$$q(k) = \frac{1}{n} \sum_i x_i(k) \varepsilon_i, \tag{28}$$

and where

$$\hat{\Sigma}(k,:) \tag{29}$$

**ALGORITHM 1** Deep Regression Ensemble

INPUT:
   Training sample, $(y, X), y \in \mathbb{R}^n, X \in \mathbb{R}^{n \times d}$
   Validation sample, $(\tilde{y}, \tilde{X})$
   Test sample, $(\mathbf{y}, \mathbf{X})$
   Number of layers, $M+1$
   Number of random feature sets per layer, $K$, and associated parameter grid $\Gamma = (\gamma_1, ..., \gamma_K)$
   Number of random features per set, $P$
   Number of ridge parameters, $L$, associated parameter grid $\Lambda = (\lambda_1, ..., \lambda_L)$
   bias parameter $a$
OUTPUT: Out of sample forecast for the test sample: $\hat{\mathbf{y}}$
   Set $x = X \in \mathbb{R}^{n \times d'}$, $d' = d$ (likewise $\tilde{x} = \tilde{X}$ and $\mathbf{x} = \mathbf{X}$)
   **for** $m \leftarrow 1$ to $M$ **do**
     **for** $k \leftarrow 1$ to $K$ **do**
       **for** $p \leftarrow 1$ to $P$ **do**
         Draw $\omega_{k,p} \in \mathbb{R}^{d' \times 1} \sim N(0, \gamma_k I)$
         Draw $b_{k,p} \in \mathbb{R} \sim U(-a, a)$
         Construct $z_{k,p}^{(m)} = \text{ReLU}(x\omega_{k,p} + b_{k,p})$ (likewise $\tilde{z}_{k,p}^{(m)}$ and $\mathbf{z}_{k,p}^{(m)}$)
       **end for**
       Set $Z_k^{(m)} = (\lambda_{k,p}^{(m)}) \in \mathbb{R}^{n \times P}$ (likewise $\tilde{Z}_k^{(m)} = (\tilde{z}_{k,p}^{(m)})$ and $\mathbf{Z}_k^{(m)} = (\mathbf{z}_{k,p}^{(m)})$)
       **for** $l \leftarrow 1$ to $L$ **do**
         Run ridge $\hat{\beta}_{k,l}^{(m)} = \arg\min_\beta \|Z_k^{(m)}\beta - y\|^2 + \lambda_l \|\beta\|^2 \in \mathbb{R}^P$
         Set $\hat{y}_{k,l}^{(m)} = Z_k^{(m)}\hat{\beta}_{k,l}^{(m)}$ (likewise $\hat{\tilde{y}}_{k,l}^{(m)} = \tilde{Z}_k^{(m)}\hat{\beta}_{k,l}^{(m)}$ and $\hat{\mathbf{y}}_{k,l}^{(m)} = \mathbf{Z}_k^{(m)}\hat{\beta}_{k,l}^{(m)}$)
       **end for**
     **end for**
     set $x = (\hat{y}_{k,l}^{(m)}) \in \mathbb{R}^{n \times d'}$, $d' = KL$ (likewise $\tilde{x} = (\hat{\tilde{y}}_{k,l}^{(m)})$ and $\mathbf{x} = (\hat{\mathbf{y}}_{k,l}^{(m)})$)
   **end for**
   *Select optimal penalization in last layer from validation sample:*
   **for** $l \leftarrow 1$ to $L$ **do**
     Run ridge $\hat{\beta}_l = \arg\min_\beta \|x\beta - y\|^2 + \lambda_l \|\beta\|^2 \in \mathbb{R}^{d'}$
     Set $\hat{y}_l = x\hat{\beta}_l$ (likewise $\hat{\tilde{y}}_l = \tilde{x}\hat{\beta}_l$ and $\hat{\mathbf{y}}_l = \mathbf{x}\hat{\beta}_l$)
   **end for**
   Find $\lambda_* = \lambda_{l^*}$ where $l^* = \arg\min_l \|\tilde{y} - \hat{\tilde{y}}_l\|^2$
   Set $\hat{\mathbf{y}} = \hat{\mathbf{y}}_{l^*}$

is the $k$-th block row of $\hat{\Sigma}$. We start with the first term: Using the block-diagonal structure of $\Sigma$ and the fact that $q$ converges to zero in probability, as well as the fact that $\beta\beta^\top \sim \text{diag}(p(k)^{-1}b_*(k))$ (see, e.g., [51]), we get[24]

$$\begin{aligned}
&\beta^\top \Sigma \hat{\beta}(k) \\
&= \beta(k)^\top \Sigma(k)(\lambda I + \hat{\Sigma}(k))^{-1}(\hat{\Sigma}(k,:)\beta + q) \\
&\stackrel{prob}{\to} \beta(k)^\top \Sigma(k)(\lambda I + \hat{\Sigma}(k))^{-1}\hat{\Sigma}(k,:)\beta \\
&= \text{tr}\left(\Sigma(\lambda I + \hat{\Sigma})^{-1}\hat{\Sigma}(k,:)\beta\beta^\top\right) \\
&\stackrel{prob}{\to} b_*(k)p(k)^{-1}\text{tr}\left(\Sigma(k)(\lambda I + \hat{\Sigma}(k))^{-1}\hat{\Sigma}(k)\right) \\
&= b_*(k)\, p(k)^{-1}\text{tr}\left(\Sigma(k)(\lambda I + \hat{\Sigma}(k))^{-1}(\lambda I + \hat{\Sigma}(k) - \lambda I)\right) \\
&\approx b_*(k)(\sigma_{*,1}(k) - \lambda \xi_k(\lambda)) = b_*(k)\nu_k(\lambda)
\end{aligned} \qquad (30)$$

---

[24] Standard martingale arguments (see, e.g., [56]) imply that, asymptotically, the traces of the resolvent can be replaced by their expectations.

by (12)-(13). We now deal with the last term in (26). It is given by

$$
\begin{aligned}
&\operatorname{tr}(\Sigma(k)\hat{\beta}(k)\hat{\beta}(k)^\top) \\
&= \operatorname{tr}(\Sigma(k)(\lambda I + \hat{\Sigma}(k))^{-1}(\hat{\Sigma}(k,:)\beta + q(k))(\hat{\Sigma}(k,:)\beta + q(k))^\top(\lambda I + \hat{\Sigma}(k))^{-1}) \\
&= \operatorname{tr}(\Sigma(k)(\lambda I + \hat{\Sigma}(k))^{-1}(\hat{\Sigma}(k,:)\beta + q(k))(\beta^\top\hat{\Sigma}(k,:)^\top + q(k)^\top)(\lambda I + \hat{\Sigma})^{-1}) \\
&\approx \operatorname{tr}(\Sigma(\lambda I + \hat{\Sigma}(k))^{-1}(\hat{\Sigma}(k,:)\beta\beta^\top\hat{\Sigma}(k,:)^\top + q(k)q(k)^\top)(\lambda I + \hat{\Sigma})^{-1})
\end{aligned}
\tag{31}
$$

where we have used the fact that the terms that are linear in $q(k)$ converge to zero in probability. Now,

$$
\begin{aligned}
E[q(k)q(k)^\top|X] &= \frac{1}{n^2}E[\sum_i x_i(k)\varepsilon_i \sum_{i_1}\varepsilon_{i_1}x_{i_1}(k)|X] \\
&= \frac{1}{n^2}E[\sum_{i,i_1} x_i(k)\varepsilon_i\varepsilon_{i_1}x_{i_1}(k)|X] \\
&= \frac{1}{n^2}E[\sum_i x_i(k)\varepsilon_i\varepsilon_i x_i(k)^\top|X] \\
&= \frac{1}{n^2}\sum_i x_i(k)x_i(k)^\top \\
&= \frac{1}{n}\hat{\Sigma}(k)
\end{aligned}
\tag{32}
$$

and it is straightforward to show that

$$
\operatorname{tr}(\Sigma(\lambda I + \hat{\Sigma}(k))^{-1}q(k)q(k)^\top(\lambda I + \hat{\Sigma})^{-1}) \stackrel{prob}{\to} \operatorname{tr}(\Sigma(\lambda I + \hat{\Sigma}(k))^{-1}\frac{1}{n}\hat{\Sigma}(k)(\lambda I + \hat{\Sigma})^{-1})
$$

converge to zero in probability. Therefore, standard martingale arguments (see, e.g., [13] and [51]) combined with the fact that $E[\beta\beta^\top] = \operatorname{diag}(p(k)^{-1}b_*(k))$ imply that

$$
\begin{aligned}
&\operatorname{tr}(\Sigma(k)\hat{\beta}(k)\hat{\beta}(k)^\top) \\
&\stackrel{prob}{\to} \operatorname{tr}(\Sigma(k)(\lambda I + \hat{\Sigma}(k))^{-1}(\hat{\Sigma}(k,:)\beta\beta^\top\hat{\Sigma}(k,:)^\top + q(k)q(k)^\top)(\lambda I + \hat{\Sigma})^{-1}) \\
&= \operatorname{tr}(\Sigma(k)(\lambda I + \hat{\Sigma}(k))^{-1}(\hat{\Sigma}(k,:)\beta\beta^\top\hat{\Sigma}(k,:)^\top + \frac{1}{n}\hat{\Sigma}(k))(\lambda I + \hat{\Sigma}(k))^{-1}) \\
&= \operatorname{tr}(\Sigma(k)(\lambda I + \hat{\Sigma}(k))^{-1}\hat{\Sigma}(k,:)\beta\beta^\top\hat{\Sigma}(k,:)^\top(\lambda I + \hat{\Sigma}(k))^{-1}) \\
&\quad + \operatorname{tr}(\Sigma(k)(\lambda I + \hat{\Sigma}(k))^{-1}\frac{1}{n}\hat{\Sigma}(k)(\lambda I + \hat{\Sigma}(k))^{-1}) \\
&\stackrel{prob}{\to} \operatorname{tr}(\Sigma(k)(\lambda I + \hat{\Sigma}(k))^{-1}\sum_{k_1}p(k_1)^{-1}b_*(k_1)\hat{\Sigma}(k,k_1)\hat{\Sigma}(k,k_1)^\top(\lambda I + \hat{\Sigma}(k))^{-1}) \\
&\quad + \operatorname{tr}(\Sigma(k)(\lambda I + \hat{\Sigma}(k))^{-1}\frac{1}{n}(\lambda I + \hat{\Sigma}(k) - \lambda I)(\lambda I + \hat{\Sigma}(k))^{-1}) \\
&= \operatorname{tr}(\Sigma(k)(\lambda I + \hat{\Sigma}(k))^{-1}p(k)^{-1}b_*(k)\hat{\Sigma}(k)^2(\lambda I + \hat{\Sigma}(k))^{-1}) \\
&\quad + \operatorname{tr}(\Sigma(k)(\lambda I + \hat{\Sigma}(k))^{-1}\sum_{k_1\neq k}p(k_1)^{-1}b_*(k_1)\hat{\Sigma}(k,k_1)\hat{\Sigma}(k,k_1)^\top(\lambda I + \hat{\Sigma}(k))^{-1}) \\
&\quad + \frac{1}{n}\operatorname{tr}(\Sigma(k)(\lambda I + \hat{\Sigma}(k))^{-1}) - \lambda\frac{1}{n}\operatorname{tr}(\Sigma(k)(\lambda I + \hat{\Sigma}(k))^{-2}) \\
&\stackrel{prob}{\to} \operatorname{tr}(\Sigma(k)(\lambda I + \hat{\Sigma}(k))^{-1}p(k)^{-1}b_*(k)\hat{\Sigma}(k)^2(\lambda I + \hat{\Sigma}(k))^{-1}) \\
&\quad + \operatorname{tr}(\Sigma(k)(\lambda I + \hat{\Sigma}(k))^{-1}\sum_{k_1\neq k}p(k_1)^{-1}b_*(k_1)\hat{\Sigma}(k,k_1)\hat{\Sigma}(k,k_1)^\top(\lambda I + \hat{\Sigma}(k))^{-1}) \\
&\quad + \xi_k(\lambda) + \lambda\xi_k'(\lambda).
\end{aligned}
\tag{33}
$$

Standard martingale arguments imply that we can replace these expressions with their expectations (in the sense of convergence in probability; see, e.g., [13] and [51]). The same calculations as above imply that

$$
\operatorname{tr}(\Sigma(k)(\lambda I + \hat{\Sigma}(k))^{-1}p(k)^{-1}b_*(k)\hat{\Sigma}(k)^2(\lambda I + \hat{\Sigma}(k))^{-1}) \to b_*(k)(\sigma_{*,1}(k) - 2\lambda c(k)^{-1}\xi_k(\lambda) - \lambda^2 c(k)^{-1}\xi_k'(\lambda)).
\tag{34}
$$

Thus, it remains to deal with the second term in (33). We have by symmetry across $i$ that

$$\mathrm{tr}(\Sigma(k)(\lambda I+\hat{\Sigma}(k))^{-1}p(k_1)^{-1}b_*(k_1)\hat{\Sigma}(k,k_1)\hat{\Sigma}(k,k_1)^\top(\lambda I+\hat{\Sigma}(k))^{-1})$$
$$= p(k_1)^{-1}b_*(k_1)\frac{1}{n^2}\,\mathrm{tr}\,E[\Sigma(k)(\lambda I+\hat{\Sigma}(k))^{-1}\bigg(nx_{i_1}(k)x_{i_1}(k_1)^\top x_{i_1}(k_1)x_{i_1}(k)^\top$$
$$+ n(n-1)\,x_{i_1}(k)x_{i_1}(k_1)^\top x_{i_2}(k_1)x_{i_2}(k)^\top\bigg)(\lambda I+\hat{\Sigma}(k))^{-1}]$$
$$= Term1 + Term2. \tag{35}$$

for some $i_1 \neq i_2$. Here,

$$Term1 = p(k_1)^{-1}b_*(k_1)\frac{1}{n}\,\mathrm{tr}\,E[\Sigma(k)(\lambda I+\hat{\Sigma}(k))^{-1}x_{i_1}(k)x_{i_1}(k_1)^\top x_{i_1}(k_1)x_{i_1}(k)^\top(\lambda I+\hat{\Sigma}(k))^{-1}] \tag{36}$$

and

$$Term2 = (1-p(k_1)^{-1})p(k_1)^{-1}b_*(k_1)\,\mathrm{tr}\,E[\Sigma(k)(\lambda I+\hat{\Sigma}(k))^{-1}x_{i_1}(k)x_{i_1}(k_1)^\top x_{i_2}(k_1)x_{i_2}(k)^\top(\lambda I+\hat{\Sigma}(k))^{-1}]. \tag{37}$$

Using the Sherman-Morrison formula (see [57]),

$$(\lambda I+\hat{\Sigma})^{-1} = (\lambda I+\hat{\Sigma}_i)^{-1} - \frac{1}{T}(\lambda I+\hat{\Sigma}_i)^{-1}x_ix_i^\top(\lambda I+\hat{\Sigma}_i)^{-1}\frac{1}{1+n^{-1}x_i^\top(\lambda I+\hat{\Sigma}_i)^{-1}x_i} \tag{38}$$

where $\hat{\Sigma}_i = \frac{1}{n}\sum_{j\neq i} x_j x_j^\top$, and defining $C = n^{-1}x_i^\top(\lambda I+\hat{\Sigma}(k)_i)^{-1}x_i$, we get

$$(\lambda I+\hat{\Sigma}(k))^{-1}x_{i_1}(k) = (\lambda I+\hat{\Sigma}(k)_i)^{-1}x_{i_1}(k)(1+C)^{-1}, \tag{39}$$

and therefore

$$Term1$$
$$= p(k_1)^{-1}b_*(k_1)\frac{1}{n}\,\mathrm{tr}\,E[\Sigma(k)(\lambda I+\hat{\Sigma}(k)_i)^{-1}x_{i_1}(k)(1+C)^{-1}$$
$$\times x_{i_1}(k_1)^\top x_{i_1}(k_1)(1+C)^{-1}x_{i_1}(k)^\top(\lambda I+\hat{\Sigma}(k)_i)^{-1}] \tag{40}$$
$$= p(k_1)^{-1}b_*(k_1)\frac{1}{n}\,\mathrm{tr}\,E[x_{i_1}(k)^\top(\lambda I+\hat{\Sigma}(k)_i)^{-1}\Sigma(k)(\lambda I+\hat{\Sigma}(k)_i)^{-1}x_{i_1}(k)x_{i_1}(k_1)^\top x_{i_1}(k_1)(1+C)^{-2}]$$

We will use the following two lemmas

**Lemma 1** *For any sequence of bounded matrices $A_P$, we have*

$$P^{-1}x_i'A_P x_i - P^{-1}\,\mathrm{tr}(A_P\Sigma) \to 0 \tag{41}$$

*is probability.*

**Proof of Lemma 1.** The proof follows directly from Proposition 2.1 in [58].

□

**Lemma 2** *We have*

$$P^{-1}\,\mathrm{tr}(Q_P((\lambda I+\hat{\Sigma}))^{-1}) - E[P^{-1}\,\mathrm{tr}(Q_P((\lambda I+\hat{\Sigma}))^{-1})] \to 0 \tag{42}$$

*almost surely for any sequence of uniformly bounded matrices $Q_P$.*

See, e.g., [51] for a proof.

Now, Lemmas 1, and 2, and the Vitali Theorem imply that

$$\frac{1}{n}x_{i_1}(k)^\top(\lambda I+\hat{\Sigma}(k)_i)^{-1}\Sigma(k)(\lambda I+\hat{\Sigma}(k)_i)^{-1}x_{i_1}(k) \to \frac{1}{n}E[\mathrm{tr}(\Sigma(k)(\lambda I+\hat{\Sigma}(k)_i)^{-1}\Sigma(k)(\lambda I+\hat{\Sigma}(k)_i)^{-1})] \tag{43}$$

in $L_2$, whereas

$$p(k_1)^{-1}x_{i_1}(k_1)^\top x_{i_1}(k_1)(1+C)^{-2} \to \sigma_*(k_1)/(1+\xi_k(\lambda))^2$$

in $L_2$. Therefore,
$$Term1 \to b_*(k_1)\widehat{\xi}_k(\lambda)\sigma_{*,1}(k_1)/(1+\xi_k(\lambda))^2 \tag{44}$$

where we have defined[25]
$$\widehat{\xi}_k(\lambda) = \frac{1}{n}E[\text{tr}(\Sigma(k)(\lambda I + \hat{\Sigma}(k))^{-1}\Sigma(k)(\lambda I + \hat{\Sigma}(k))^{-1})]$$

We will now need the following lemma.

**Lemma 3** *We have*
$$\frac{1}{n}E[\text{tr}(\Sigma(k)(\lambda I + \hat{\Sigma}(k))^{-1}\Sigma(k)(\lambda I + \hat{\Sigma}(k))^{-1})] \to \widehat{\xi}_k(\lambda) = (\xi_k(\lambda) + \lambda\xi'_k(\lambda))(1+\xi_k(\lambda))^2 \tag{45}$$

**Proof of Lemma 3.** We have
$$\frac{1}{n}\text{ tr } E[\Sigma(k)(\lambda I + \hat{\Sigma}(k)_i)^{-1}] \to \xi_k(\lambda)$$

by (13) and therefore
$$\frac{1}{n}\text{ tr } E[(\lambda I + \hat{\Sigma}(k)_i)^{-1}\Sigma(k)(\lambda I + \hat{\Sigma}(k)_i)^{-1}] = \frac{1}{n}\text{ tr } E[\Sigma(k)(\lambda I + \hat{\Sigma}(k)_i)^{-2}] \to -\xi'_k(\lambda)\,.$$

Lemmas 1, and 2, and the Vitali Theorem together with the fact that $x_{i_1}$ is independent of $\hat{\Sigma}(k)_i$ imply that
$$\frac{1}{n}x_{i_1}(k)^\top(\lambda I + \hat{\Sigma}(k)_i)^{-1}\Sigma(k)(\lambda I + \hat{\Sigma}(k)_i)^{-1}x_{i_1}(k) \approx \frac{1}{n}\text{ tr } E[\Sigma(k)(\lambda I + \hat{\Sigma}(k)_i)^{-1}\Sigma(k)(\lambda I + \hat{\Sigma}(k)_i)^{-1}]$$
$$\approx \frac{1}{n}\text{ tr } E[\Sigma(k)(\lambda I + \hat{\Sigma}(k))^{-1}\Sigma(k)(\lambda I + \hat{\Sigma}(k)_i)^{-1}] \tag{46}$$

is probability. In the next equation, to simplify the expressions, we will use $X \approx Y$ to denote the fact that $X - Y \to 0$ as $n \to \infty$. By (39) and (46),

$$\begin{aligned}
\xi_k(\lambda) &\approx \frac{1}{n}\text{ tr } E[\Sigma(k)(\lambda I + \hat{\Sigma}(k))^{-1}] \\
&= \frac{1}{n}\text{ tr } E[(\lambda I + \hat{\Sigma}(k))(\lambda I + \hat{\Sigma}(k))^{-1}\Sigma(k)(\lambda I + \hat{\Sigma}(k))^{-1}] \\
&\approx -\lambda\xi'_k(\lambda) + \frac{1}{n}\text{ tr } E[\hat{\Sigma}(k)(\lambda I + \hat{\Sigma}(k))^{-1}\Sigma(k)(\lambda I + \hat{\Sigma}(k))^{-1}] \\
&= \{\hat{\Sigma}(k) = n^{-1}\sum_i x_i(k)x_i(k)^\top\} \\
&= -\lambda\xi'_k(\lambda) + \frac{1}{n^2}\sum_i \text{tr } E[x_i(k)x_i(k)^\top(\lambda I + \hat{\Sigma}(k))^{-1}\Sigma(k)(\lambda I + \hat{\Sigma}(k))^{-1}] \\
&= -\lambda\xi'_k(\lambda) + \frac{1}{n}\text{ tr } E[x_i(k)x_i(k)^\top(\lambda I + \hat{\Sigma}(k))^{-1}\Sigma(k)(\lambda I + \hat{\Sigma}(k))^{-1}] \\
&= -\lambda\xi'_k(\lambda) + \frac{1}{n}\text{ tr } E[(\lambda I + \hat{\Sigma}(k))^{-1}x_i(k)x_i(k)^\top(\lambda I + \hat{\Sigma}(k))^{-1}\Sigma(k)] \\
&= -\lambda\xi'_k(\lambda) \\
&\quad + \frac{1}{n}\text{ tr } E[(\lambda I + \hat{\Sigma}(k)_i)^{-1}x_i(k)(1+C)^{-1}x_i(k)^\top(\lambda I + \hat{\Sigma}(k))^{-1}\Sigma(k)] \\
&= -\lambda\xi'_k(\lambda) \\
&\quad + \frac{1}{n}\text{ tr } E[(1+C)^{-2}x_i(k)^\top(\lambda I + \hat{\Sigma}(k)_i)^{-1}\Sigma(k)(\lambda I + \hat{\Sigma}(k)_i)^{-1}x_i(k)] \\
&\approx -\lambda\xi'_k(\lambda) + (1+\xi_k(\lambda))^{-2}\widehat{\xi}_k(\lambda)
\end{aligned} \tag{47}$$

and the claim follows. The proof of Lemma 3 is complete. $\square$

---

[25] We systematically abuse the notation and use $\widehat{\xi}_k(\lambda)$ to denote both finite $n$ object and its $n \to \infty$ limit.

Thus, it remains to deal with $Term2$ in (35). By (39),

$$
\begin{aligned}
Term2 &\approx p(k_1)^{-1} b_*(k_1) \operatorname{tr} E[\Sigma(k)(\lambda I + \hat{\Sigma}(k))^{-1} x_{i_1}(k) x_{i_1}(k_1)^\top x_{i_2}(k_1) x_{i_2}(k)^\top (\lambda I + \hat{\Sigma}(k))^{-1}] \\
&= p(k_1)^{-1} b_*(k_1) \operatorname{tr} E[x_{i_2}(k)^\top (\lambda I + \hat{\Sigma}(k))^{-1} \Sigma(k)(\lambda I + \hat{\Sigma}(k))^{-1} x_{i_1}(k) x_{i_1}(k_1)^\top x_{i_2}(k_1)] \\
&\approx p(k_1)^{-1} b_*(k_1) \operatorname{tr} E[(1+C)^{-1} x_{i_2}(k)^\top (\lambda I + \hat{\Sigma}(k)_{i_2})^{-1} \\
&\quad \times \Sigma(k)(\lambda I + \hat{\Sigma}(k)_{i_1})^{-1} x_{i_1}(k)(1+C)^{-1} x_{i_1}(k_1)^\top x_{i_2}(k_1)] \\
&\approx p(k_1)^{-1} b_*(k_1) \operatorname{tr} E[(1+C)^{-1} x_{i_2}(k)^\top \\
&\quad \times \left( (\lambda I + \hat{\Sigma}(k)_{i_1,i_2})^{-1} - \frac{1}{n}(\lambda I + \hat{\Sigma}(k)_{i_1,i_2})^{-1} x_{i_1}(k)(1+C)^{-1} x_{i_1}(k)^\top (\lambda I + \hat{\Sigma}(k)_{i_1,i_2})^{-1} \right) \\
&\quad \times \Sigma(k) \left( (\lambda I + \hat{\Sigma}(k)_{i_1,i_2})^{-1} - \frac{1}{n}(\lambda I + \hat{\Sigma}(k)_{i_1,i_2})^{-1} x_{i_2}(k)(1+C)^{-1} x_{i_2}(k)^\top (\lambda I + \hat{\Sigma}(k)_{i_1,i_2})^{-1} \right) \\
&\quad x_{i_1}(k)(1+C)^{-1} x_{i_1}(k_1)^\top x_{i_2}(k_1)] \\
&= p(k_1)^{-1} b_*(k_1) \operatorname{tr} E[(1+C)^{-1} x_{i_2}(k) \\
&\quad \times \left( (\lambda I + \hat{\Sigma}(k)_{i_1,i_2})^{-1} - \frac{1}{n}(\lambda I + \hat{\Sigma}(k)_{i_1,i_2})^{-1} x_{i_1}(k)(1+C)^{-1} x_{i_1}(k)^\top (\lambda I + \hat{\Sigma}(k)_{i_1,i_2})^{-1} \right) \\
&\quad \times \Sigma(k) \left( (\lambda I + \hat{\Sigma}(k)_{i_1,i_2})^{-1} - \frac{1}{n}(\lambda I + \hat{\Sigma}(k)_{i_1,i_2})^{-1} x_{i_2}(k)(1+C)^{-1} x_{i_2}(k)^\top (\lambda I + \hat{\Sigma}(k)_{i_1,i_2})^{-1} \right) \\
&\quad x_{i_1}(k)(1+C)^{-1} x_{i_1}(k_1)^\top x_{i_2}(k_1)] \\
&= p(k_1)^{-1} b_*(k_1) \operatorname{tr} E[(1+C)^{-1} x_{i_2}(k)^\top \\
&\quad \times \Bigg( (\lambda I + \hat{\Sigma}(k)_{i_1,i_2})^{-1} \Sigma(k)(\lambda I + \hat{\Sigma}(k)_{i_1,i_2})^{-1} \\
&\quad - \frac{1}{n}(\lambda I + \hat{\Sigma}(k)_{i_1,i_2})^{-1} x_{i_1}(k)(1+C)^{-1} x_{i_1}(k)^\top (\lambda I + \hat{\Sigma}(k)_{i_1,i_2})^{-1} \Sigma(k)(\lambda I + \hat{\Sigma}(k)_{i_1,i_2})^{-1} \\
&\quad - (\lambda I + \hat{\Sigma}(k)_{i_1,i_2})^{-1} \Sigma(k) \frac{1}{n}(\lambda I + \hat{\Sigma}(k)_{i_1,i_2})^{-1} x_{i_2}(k)(1+C)^{-1} x_{i_2}(k)^\top (\lambda I + \hat{\Sigma}(k)_{i_1,i_2})^{-1} \\
&\quad + \frac{1}{n}(\lambda I + \hat{\Sigma}(k)_{i_1,i_2})^{-1} x_{i_1}(k)(1+C)^{-1} x_{i_1}(k)^\top (\lambda I + \hat{\Sigma}(k)_{i_1,i_2})^{-1} \Sigma(k) \\
&\quad \times \frac{1}{n}(\lambda I + \hat{\Sigma}(k)_{i_1,i_2})^{-1} x_{i_2}(k)(1+C)^{-1} x_{i_2}(k)^\top (\lambda I + \hat{\Sigma}(k)_{i_1,i_2})^{-1} \Bigg) x_{i_1}(k)(1+C)^{-1} x_{i_1}(k_1)^\top x_{i_2}(k_1)] \\
&= Term21 + Term22 + Term23 + Term24 \,.
\end{aligned}
\tag{48}
$$

Note that the different $1+C$ factors differ from each other slightly, but we will abuse the notation and treat them as identical. Dealing with them separately requires minor modifications in the proofs. By direct calculation,

$$E[x_{i_2}(k)^\top Q x_{i_2}(k_1) | x_{i_2}] = \operatorname{tr}(Q \Psi_{2,1}) \tag{49}$$

for any $Q$ independent of $x_{i_2}$. Thus,

$$
\begin{aligned}
Term21 &= p(k_1)^{-1} b_*(k_1) \operatorname{tr} E[(1+C)^{-1} x_{i_2}(k)^\top \\
&\quad \times \left( (\lambda I + \hat{\Sigma}(k)_{i_1,i_2})^{-1} \Sigma(k)(\lambda I + \hat{\Sigma}(k)_{i_1,i_2})^{-1} \right) x_{i_1}(k)(1+C)^{-1} x_{i_1}(k_1)^\top x_{i_2}(k_1)] \\
&= p(k_1)^{-1} b_*(k_1) \operatorname{tr} E[(1+C)^{-2} x_{i_2}(k)^\top Q x_{i_2}(k_1)] \;=\; 0
\end{aligned}
\tag{50}
$$

because there are no cross-group correlations; here we have defined

$$Q = \left( (\lambda I + \hat{\Sigma}(k)_{i_1,i_2})^{-1} \Sigma(k)(\lambda I + \hat{\Sigma}(k)_{i_1,i_2})^{-1} \right) x_{i_1}(k) x_{i_1}(k_1)^\top \,.$$

Proceeding to the next term in (48), we get

$$
\begin{aligned}
Term22 &= p(k_1)^{-1} b_*(k_1) \operatorname{tr} E[(1+C)^{-1} x_{i_2}(k)^\top \\
&\quad \times \left( -\frac{1}{n}(\lambda I + \hat\Sigma(k)_{i_1,i_2})^{-1} x_{i_1}(k)(1+C)^{-1} x_{i_1}(k)^\top (\lambda I + \hat\Sigma(k)_{i_1,i_2})^{-1} \Sigma(k)(\lambda I + \hat\Sigma(k)_{i_1,i_2})^{-1} \right) \\
&\quad x_{i_1}(k)(1+C)^{-1} x_{i_1}(k_1)^\top x_{i_2}(k_1)]
\end{aligned}
\tag{51}
$$

We have

$$
\begin{aligned}
&\frac{1}{n} x_{i_1}(k)^\top (\lambda I + \hat\Sigma(k)_{i_1,i_2})^{-1} \Sigma(k)(\lambda I + \hat\Sigma(k)_{i_1,i_2})^{-1} x_{i_1}(k) \\
&\to \frac{1}{n} \operatorname{tr} E[\Sigma(k)(\lambda I + \hat\Sigma(k)_{i_1,i_2})^{-1} \Sigma(k)(\lambda I + \hat\Sigma(k)_{i_1,i_2})^{-1}] \\
&= \widehat{\xi}_k(\lambda)
\end{aligned}
\tag{52}
$$

is probability by Lemmas 1 and 2 and the Vitali Theorem. Hence,

$$
\begin{aligned}
Term22 &\to p(k_1)^{-1} b_*(k_1) \operatorname{tr} E[(1+C)^{-1} x_{i_2}(k) \\
&\quad \times \left( -(\lambda I + \hat\Sigma(k)_{i_1,i_2})^{-1} x_{i_1}(k)^\top (1+C)^{-1} \widehat{\xi}_k(\lambda) \right) (1+C)^{-1} x_{i_1}(k_1) x_{i_2}(k_1)^\top] \\
&\to 0
\end{aligned}
\tag{53}
$$

where we have used simple Cauchy-Schwarz-based arguments to pass to an $L_2$-limit.[26] Once again, the limit is zero due to the assumed zero cross-group covariance. Proceeding to the next term in (48), we get

$$
\begin{aligned}
Term23 &\approx p(k_1)^{-1} b_*(k_1) E[(1+C)^{-1} x_{i_2}(k)^\top \\
&\quad \left( -(\lambda I + \hat\Sigma(k)_{i_1,i_2})^{-1} \Sigma(k) \frac{1}{n}(\lambda I + \hat\Sigma(k)_{i_1,i_2})^{-1} x_{i_2}(k)(1+C)^{-1} x_{i_2}(k)^\top (\lambda I + \hat\Sigma(k)_{i_1,i_2})^{-1} \right) \\
&\quad x_{i_1}(k)(1+C)^{-1} x_{i_1}(k_1)^\top x_{i_2}(k_1)] \\
&= -b_*(k_1) E[XY]
\end{aligned}
\tag{54}
$$

where we have defined

$$
X = -(1+C)^{-1} x_{i_2}(k)^\top (\lambda I + \hat\Sigma(k)_{i_1,i_2})^{-1} \Sigma(k) \frac{1}{n}(\lambda I + \hat\Sigma(k)_{i_1,i_2})^{-1} x_{i_2}(k)
$$

and

$$
Y = p(k_1)^{-1}(1+C)^{-2} x_{i_1}(k_1)^\top x_{i_2}(k_1) x_{i_2}(k)^\top (\lambda I + \hat\Sigma(k)_{i_1,i_2})^{-1} x_{i_1}(k).
$$

By Lemma 3 and (13), $X \to (1+\xi_k(\lambda))^{-1} \widehat{\xi}(\lambda)$ in $L_2$, whereas $Y$ has a bounded $L_2$-norm. Then,

$$
E[XY] - (1+\xi_k(\lambda))^{-1} \widehat{\xi}(\lambda) E[Y] \to 0
$$

Integrating over $x_{i_2}$ gives

$$
E[Y] = 0
$$

Thus, $Term23$ in (48) converges to zero due to zero cross-correlation across signal groups.

Finally, the last term in (48) also converges to zero by a similar argument. See [51] for details.

Gathering the terms, we get

$$
\begin{aligned}
&\operatorname{tr}(\Sigma(k)\hat\beta(k)\hat\beta(k)') \\
&\overset{prob}{\to} b_*(k)(\sigma_{*,1}(k) - 2\lambda c(k)^{-1}\xi_k(\lambda) - \lambda^2 c(k)^{-1}\xi'_k(\lambda)) + \xi_k(\lambda) + \lambda \xi'_k(\lambda) \\
&\quad + \operatorname{tr}(\Sigma(k)(\lambda I + \hat\Sigma(k))^{-1} \sum_{k_1 \neq k} p(k_1)^{-1} b_*(k_1) \hat\Sigma(k,k_1) \hat\Sigma(k,k_1)^\top (\lambda I + \hat\Sigma(k))^{-1}) \\
&\overset{prob}{\to} b_*(k)(\sigma_{*,1}(k) - 2\lambda c(k)^{-1}\xi_k(\lambda) - \lambda^2 c(k)^{-1}\xi'_k(\lambda)) + (\xi_k(\lambda) + \lambda \xi'_k(\lambda))(1 + \sum_{k_1 \neq k} b_*(k_1)\sigma_{*,1}(k_1)) \\
&= b_*(k)(\sigma_{*,1}(k) - 2\lambda c(k)^{-1}\xi_k(\lambda) - \lambda^2 c(k)^{-1}\xi'_k(\lambda)) + (\xi_k(\lambda) + \lambda \xi'_k(\lambda))(1 + \bar b_* - b_*(k)\sigma_{*,1}(k)).
\end{aligned}
\tag{55}
$$

---

[26]Note that it may seem that we need six bounded moments for the signals. But, in fact, the normalization by $1+C$ ensures all the necessary terms stay bounded.

We now proceed to computing the risk of $\beta$ estimated using the full model. Let $\Pi_k$ be the orthogonal projection onto the $k$-th component of the decomposition $\mathbb{R}^P = \oplus_{k=1}^K \mathbb{R}^{p(k)}$. Then,

$$\begin{aligned}
&\beta^\top \Sigma \hat{\beta} \\
&= \beta^\top \Sigma (\lambda I + \hat{\Sigma})^{-1}(\hat{\Sigma}\beta + q) \\
&\stackrel{prob}{\to} \beta^\top \Sigma (\lambda I + \hat{\Sigma})^{-1}\hat{\Sigma}\beta \\
&= \operatorname{tr}\left(\Sigma(\lambda I + \hat{\Sigma})^{-1}\hat{\Sigma}\beta\beta^\top\right) \\
&\stackrel{prob}{\to} \sum_k b_*(k)p(k)^{-1} \operatorname{tr}\left(\Sigma(\lambda I + \hat{\Sigma})^{-1}\hat{\Sigma}\Pi_k\right) \\
&= \sum_k b_*(k)p(k)^{-1} \operatorname{tr}\left(\Sigma(\lambda I + \hat{\Sigma})^{-1}(\lambda I + \hat{\Sigma} - \lambda I)\Pi_k\right)
\end{aligned} \tag{56}$$

Under the assumption of identical $p(k) = p$ and identical eigenvalue distributions, $\operatorname{tr}\left(\Sigma(\lambda I + \hat{\Sigma})^{-1}(\lambda I + \hat{\Sigma} - \lambda I)\Pi_k\right)$ is independent of $k$ and, hence,

$$p(k)^{-1} \operatorname{tr}\left(\Sigma(\lambda I + \hat{\Sigma})^{-1}(\lambda I + \hat{\Sigma} - \lambda I)\Pi_k\right) = P^{-1} \operatorname{tr}\left(\Sigma(\lambda I + \hat{\Sigma})^{-1}(\lambda I + \hat{\Sigma} - \lambda I)\right)$$

where $p(k) = p$ and $P = pK$. Thus, the same arguments as above imply that

$$\beta^\top \Sigma \hat{\beta} = \sum_k b_*(k)\nu(\lambda)$$

Similarly, using the same arguments as in (33), we get

$$\begin{aligned}
&\operatorname{tr}(\Sigma\hat{\beta}\hat{\beta}^\top) \\
&\stackrel{prob}{\to} \operatorname{tr}(\Sigma(\lambda I + \hat{\Sigma})^{-1}(\hat{\Sigma}\beta\beta^\top\hat{\Sigma}^\top + q(k)q(k)^\top)(\lambda I + \hat{\Sigma})^{-1}) \\
&= \operatorname{tr}(\Sigma(\lambda I + \hat{\Sigma})^{-1}(\hat{\Sigma}\beta\beta^\top\hat{\Sigma}^\top + \frac{1}{n}\hat{\Sigma})(\lambda I + \hat{\Sigma})^{-1}) \\
&= \operatorname{tr}(\Sigma(\lambda I + \hat{\Sigma})^{-1}\hat{\Sigma}\beta\beta^\top\hat{\Sigma}^\top(\lambda I + \hat{\Sigma})^{-1}) \\
&\quad + \operatorname{tr}(\Sigma(\lambda I + \hat{\Sigma})^{-1}\frac{1}{n}\hat{\Sigma}(\lambda I + \hat{\Sigma})^{-1}) \\
&\stackrel{prob}{\to} \sum_k b_*(k)p(k)^{-1} \operatorname{tr}(\Sigma(\lambda I + \hat{\Sigma})^{-1}\hat{\Sigma}\Pi_k\hat{\Sigma}(\lambda I + \hat{\Sigma})^{-1}) \\
&\quad + \operatorname{tr}(\Sigma(\lambda I + \hat{\Sigma})^{-1}\frac{1}{n}(\lambda I + \hat{\Sigma} - \lambda I)(\lambda I + \hat{\Sigma})^{-1}) \\
&= \sum_k b_*(k)p(k)^{-1} \operatorname{tr}(\Sigma(\lambda I + \hat{\Sigma})^{-1}\hat{\Sigma}\Pi_k\hat{\Sigma}(\lambda I + \hat{\Sigma})^{-1}) \\
&\quad + \frac{1}{n}\operatorname{tr}(\Sigma(\lambda I + \hat{\Sigma})^{-1}) - \lambda\frac{1}{n}\operatorname{tr}(\Sigma(\lambda I + \hat{\Sigma})^{-2}) \\
&\stackrel{prob}{\to} \sum_k b_*(k)p(k)^{-1} \operatorname{tr}(\Sigma(\lambda I + \hat{\Sigma})^{-1}\hat{\Sigma}\Pi_k\hat{\Sigma}(\lambda I + \hat{\Sigma})^{-1}) \\
&\quad + \xi(\lambda) + \lambda\xi'(\lambda).
\end{aligned} \tag{57}$$

Under the assumption of identical $p(k) = p$ and identical eigenvalue distributions, we get that

$$p(k)^{-1} \operatorname{tr}(\Sigma(\lambda I + \hat{\Sigma})^{-1}\hat{\Sigma}\Pi_k\hat{\Sigma}(\lambda I + \hat{\Sigma})^{-1}) = P^{-1}\operatorname{tr}(\Sigma(\lambda I + \hat{\Sigma})^{-1}\hat{\Sigma}\hat{\Sigma}(\lambda I + \hat{\Sigma})^{-1})$$

and the same argument as above implies that

$$P^{-1}\operatorname{tr}(\Sigma(\lambda I + \hat{\Sigma})^{-1}\hat{\Sigma}\hat{\Sigma}(\lambda I + \hat{\Sigma})^{-1}) \stackrel{prob}{\to} \sigma_* - 2\lambda c^{-1}\xi(\lambda) - \lambda^2 c^{-1}\xi'(\lambda) = \hat{\nu}(\lambda) \tag{58}$$

The proof is complete. $\square$

**Proof of Proposition 4.** By direct calculation,
$$n^{-1} \operatorname{tr} E[(\lambda_1 I + \hat{\Sigma})^{-1}(\lambda_2 I + \hat{\Sigma})^{-1}] = n^{-1}(\lambda_1 - \lambda_2)^{-1}(m(-\lambda_2; c) - m(-\lambda_1; c)) \tag{59}$$

We then have

$$\begin{aligned}
&\operatorname{tr}\left((\lambda_1 I + \hat{\Sigma})^{-1}(\hat{\Sigma}\beta + q)(\hat{\Sigma}\beta + q)^\top (\lambda_2 I + \hat{\Sigma})^{-1}\Sigma\right) \\
&\approx \operatorname{tr}\left(\Sigma(\Sigma(\lambda_1 I + \hat{\Sigma})^{-1}(\hat{\Sigma}\beta\beta^\top\hat{\Sigma} + qq')(\lambda_2 I + \hat{\Sigma})^{-1})\right) \\
&\approx \operatorname{tr}(\Sigma(\lambda_1 I + \hat{\Sigma})^{-1}(\hat{\Sigma}\beta\beta^\top\hat{\Sigma} + \tfrac{1}{n}\hat{\Sigma})(\lambda_2 I + \hat{\Sigma})^{-1}) \\
&= \operatorname{tr}(\Sigma(\lambda_1 I + \hat{\Sigma})^{-1}\hat{\Sigma}\beta\beta^\top\hat{\Sigma}(\lambda_2 I + \hat{\Sigma})^{-1}) \\
&\quad + \operatorname{tr}(\Sigma(\lambda_1 I + \hat{\Sigma})^{-1}\tfrac{1}{n}\hat{\Sigma}(\lambda_2 I + \hat{\Sigma})^{-1}) \\
&= \operatorname{tr}(\hat{\Sigma}(\lambda_1 I + \hat{\Sigma})^{-1}\Sigma(\lambda_2 I + \hat{\Sigma})^{-1}\hat{\Sigma}\beta\beta^\top) \\
&\quad + \operatorname{tr}(\Sigma(\lambda_1 I + \hat{\Sigma})^{-1}\tfrac{1}{n}(\lambda_2 I + \hat{\Sigma} - \lambda_2 I)(\lambda_2 I + \hat{\Sigma})^{-1}) \\
&\xrightarrow{prob} b_* P^{-1} \operatorname{tr}[\hat{\Sigma}(\lambda_1 I + \hat{\Sigma})^{-1}\Sigma(\lambda_2 I + \hat{\Sigma})^{-1}\hat{\Sigma}] \\
&\quad + \tfrac{1}{n}\operatorname{tr}(\Sigma[(\lambda_1 I + \hat{\Sigma})^{-1}(\lambda_2 I + \hat{\Sigma})(\lambda_2 I + \hat{\Sigma})^{-1}]) \\
&\quad - \lambda_2 \tfrac{1}{n}\operatorname{tr}(\Sigma E[(\lambda_1 I + \hat{\Sigma})^{-1}(\lambda_2 I + \hat{\Sigma})^{-1}]) \\
&= b_* P^{-1} \operatorname{tr} E[\Sigma\hat{\Sigma}(\lambda_1 I + \hat{\Sigma})^{-1}(\lambda_2 I + \hat{\Sigma})^{-1}] \\
&\quad + \tfrac{1}{n}\operatorname{tr}(\Sigma E[(\lambda_1 I + \hat{\Sigma})^{-1}]) \\
&\quad - \lambda_2 \tfrac{1}{n}\operatorname{tr}(\Sigma E[(\lambda_2 I + \hat{\Sigma})^{-1}(\lambda_1 I + \hat{\Sigma})^{-1}]) \\
&= Term1 + Term2 + Term3\,.
\end{aligned} \tag{60}$$

We now proceed with each term:
$$\begin{aligned}
&(\lambda_1 I + \hat{\Sigma})^{-1}\hat{\Sigma}^2(\lambda_2 I + \hat{\Sigma})^{-1} \\
&= \{\text{all matrices commute}\} = (\lambda_2 - \lambda_1)^{-1}((\lambda_1 I + \hat{\Sigma})^{-1} - (\lambda_2 I + \hat{\Sigma})^{-1})\hat{\Sigma}^2
\end{aligned} \tag{61}$$

whereas
$$\begin{aligned}
&(\lambda I + \hat{\Sigma})^{-1}\hat{\Sigma}^2 \\
&= (\lambda I + \hat{\Sigma})^{-1}(\hat{\Sigma}^2 + 2\lambda\hat{\Sigma} + \lambda^2 I - 2\lambda\hat{\Sigma} - \lambda^2 I) \\
&= (\lambda I + \hat{\Sigma})^{-1}(\hat{\Sigma}^2 + 2\lambda\hat{\Sigma} + \lambda^2 I - 2z(\hat{\Sigma} + \lambda I) + \lambda^2 I) \\
&= (\lambda I + \hat{\Sigma})^{-1}\left((\lambda I + \hat{\Sigma})^2 - 2z(\hat{\Sigma} + \lambda I) + \lambda^2 I\right) \\
&= (\lambda I + \hat{\Sigma}) - 2zI + \lambda^2(\lambda I + \hat{\Sigma})^{-1}\,.
\end{aligned} \tag{62}$$

Therefore,
$$\begin{aligned}
Term1 &= b_* P^{-1} \operatorname{tr} E[\Sigma\hat{\Sigma}(\lambda_1 I + \hat{\Sigma})^{-1}(\lambda_2 I + \hat{\Sigma})^{-1}] \\
&= b_* P^{-1}(\lambda_2 - \lambda_1)^{-1} \operatorname{tr} E[\Sigma\bigg(((\lambda_1 I + \hat{\Sigma}) - 2\lambda_1 I + \lambda_1^2(\lambda_1 I + \hat{\Sigma})^{-1}) \\
&\quad - ((\lambda_2 I + \hat{\Sigma}) - 2\lambda_2 I + \lambda_2^2(\lambda_2 I + \hat{\Sigma})^{-1})\bigg)] \\
&= b_* \Sigma_{*,1} + b_* c^{-1}(\lambda_2 - \lambda_1)^{-1}(\lambda_1^2 \xi(\lambda_1) - \lambda_2^2 \xi(\lambda_2))
\end{aligned} \tag{63}$$

Furthermore,
$$Term2 = \tfrac{1}{n}\operatorname{tr}(\Sigma E[(\lambda_1 I + \hat{\Sigma})^{-1}]) \to \xi(\lambda_1)$$

Finally,
$$\begin{aligned} Term3 &= -\lambda_2 \frac{1}{n} \operatorname{tr}(\Sigma E[(\lambda_2 I + \hat{\Sigma})^{-1}(\lambda_1 I + \hat{\Sigma})^{-1}]) \\ &= -\lambda_2 \frac{1}{n}(\lambda_2 - \lambda_1)^{-1} \operatorname{tr}(\Sigma E[-(\lambda_2 I + \hat{\Sigma})^{-1} + (\lambda_1 I + \hat{\Sigma})^{-1}]) \\ &\to -\lambda_2 (\lambda_2 - \lambda_1)^{-1} (\xi(\lambda_1) - \xi(\lambda_2)) \end{aligned} \qquad (64)$$

Thus,
$$\begin{aligned} \hat{\beta}_{\lambda_1}^\top \Sigma \hat{\beta}_{\lambda_2} &\overset{prob}{\to} Term1 \; + \; Term2 \; + \; Term3 \\ &\overset{prob}{\to} b_* \Sigma_{*,1} \; + \; b_* c^{-1} (\lambda_2 - \lambda_1)^{-1} (\lambda_1^2 \xi(\lambda_1) - \lambda_2^2 \xi(\lambda_2)) \\ &\quad + \; \xi(\lambda_1) - \lambda_2 (\lambda_2 - \lambda_1)^{-1} (\xi(\lambda_1) - \xi(\lambda_2)) \\ &= b_* \Big( \Sigma_{*,1} \; + \; c^{-1} (\lambda_2 - \lambda_1)^{-1} (\lambda_1^2 \xi(\lambda_1) - \lambda_2^2 \xi(\lambda_2)) \Big) \; + \; (\lambda_2 - \lambda_1)^{-1} (\lambda_2 \xi(\lambda_2) - \lambda_1 \xi(\lambda_1)) \\ &= \Gamma(\lambda_1, \lambda_2) \; = \; \Gamma(\lambda_2, \lambda_1) \end{aligned} \qquad (65)$$

$\square$

**Proof of Proposition 4.** Let $a$ be the scale of the full-ridge model. Then, we will use
$$\bar{b}_* \;=\; \sum_k b_*(k) \qquad (66)$$

We have
$$Risk(flat) \;=\; \bar{b}_* \;-\; \frac{(\bar{b}_* \nu(\lambda; cK))^2}{\bar{b}_* \hat{\nu}(\lambda; cK) \;-\; (cK)\, \nu'(\lambda; cK)} \qquad (67)$$

and
$$Risk(ensemble) \;=\; \bar{b}_* \;-\; \sum_k \frac{(b_*(k)\nu(\lambda; c))^2}{b_*(k)\, \hat{\nu}(\lambda; c) \;-\; c\,(1 + (\bar{b}_* - b_*(k)))\nu'(\lambda; c)} \qquad (68)$$

We have
$$\xi(\lambda) \;=\; \frac{1 - \lambda m(-\lambda; c)}{c^{-1} - 1 + \lambda m(-\lambda; c)} \qquad (69)$$

We will also need the following asymptotic characterization (see, e.g., [51]):
$$(1 - \lambda m(-\lambda; c)) \;=\; \lambda^{-1} \sigma_{*,1} - \lambda^{-2}(\sigma_{*,2} + c\sigma_{*,1}^2) + \lambda^{-3}(\sigma_{*,3} + 3c\sigma_{*,2}\sigma_{*,1} + c^2 \sigma_{*,1}^3) + O(\lambda^{-4}). \qquad (70)$$

Then,
$$\begin{aligned} \xi(\lambda) &= -1 + \frac{1}{1 - c + c\lambda m} \\ &= -1 + \frac{1}{1 - c(1 - \lambda m(-\lambda))} \;=\; c(1 - \lambda m(-\lambda)) + c^2(1 - \lambda m(-\lambda))^2 + c^3(1 - \lambda m(-\lambda))^3 + O(\lambda^{-4}) \\ &= c\Big(\lambda^{-1}\sigma_{*,1} - \lambda^{-2}(\sigma_{*,2} + c\sigma_{*,1}^2) + \lambda^{-3}(\sigma_{*,3} + 3c\sigma_{*,2}\sigma_{*,1} + c^2\sigma_{*,1}^3)\Big) \\ &\quad + c^2(\lambda^{-2}\sigma_{*,1}^2 - 2\lambda^{-3}\sigma_{*,1}(\sigma_{*,2} + c\sigma_{*,1}^2)) \;+\; c^3 \lambda^{-3} \sigma_{*,1}^3 \;+\; O(\lambda^{-4}) \\ &= c\lambda^{-1}\sigma_{*,1} - c\lambda^{-2}\sigma_{*,2} + c\lambda^{-3}(\sigma_{*,3} + 3c\sigma_{*,2}\sigma_{*,1} + c^2\sigma_{*,1}^3 - 2c\sigma_{*,1}\sigma_{*,2} - 2c^2\sigma_{*,1}^3 + c^2\sigma_{*,1}^3) \;+\; O(\lambda^{-4}) \\ &= c\lambda^{-1}\sigma_{*,1} - c\lambda^{-2}\sigma_{*,2} + c\lambda^{-3}(\sigma_{*,3} + c\sigma_{*,2}\sigma_{*,1}) \;+\; O(\lambda^{-4}), \end{aligned} \qquad (71)$$

and hence
$$\nu(\lambda; c) \;=\; \lambda^{-1}\sigma_{*,2} - \lambda^{-2}(\sigma_{*,3} + c\sigma_{*,2}\sigma_{*,1}) + O(\lambda^{-3})$$

and
$$\nu'(\lambda; c) \;=\; -\lambda^{-2}\sigma_{*,2} + 2\lambda^{-3}(\sigma_{*,3} + c\sigma_{*,2}\sigma_{*,1}) + O(\lambda^{-4})$$

and
$$\hat{\nu}(\lambda; c) \;=\; \nu(\lambda; c) + \lambda \nu'(\lambda; c) \;=\; \lambda^{-2}(\sigma_{*,3} + c\sigma_{*,2}\sigma_{*,1}) + O(\lambda^{-3})$$

and hence
$$\frac{\nu^2}{b_* \hat{\nu} \;-\; c\nu'} \;=\; \frac{\sigma_{*,2}^2 - 2\lambda^{-1}\sigma_{*,2}(\sigma_{*,3} + c\sigma_{*,2}\sigma_{*,1}) + \lambda^{-2}(\sigma_{*,3} + c\sigma_{*,2}\sigma_{*,1})^2 \;+\; O(\lambda^{-1})}{b_*(\sigma_{*,3} + c\sigma_{*,2}\sigma_{*,1}) \;+\; O(\lambda^{-1}) + c\sigma_{*,2}} \qquad (72)$$

Thus, for $\lambda \to \infty$, the desired inequality is equivalent to

$$\sum_k \frac{b_*(k)^2 \sigma_{*,2}^2}{b_*(k)(\sigma_{*,3} + c\sigma_{*,2}\sigma_{*,1}) + c(1 + (\bar{b}_* - b_*(k))\sigma_{*,1})\sigma_{*,2}} > \frac{(\bar{b}_*(k))^2 \sigma_{*,2}^2}{\bar{b}_*(k)(\sigma_{*,3} + Kc\sigma_{*,2}\sigma_{*,1}) + Kc\sigma_{*,2}}, \quad (73)$$

which is in turn equivalent to

$$\sum_k \frac{b_*(k)^2 \sigma_{*,2}^2}{b_*(k)\sigma_{*,3} + c(1 + \bar{b}_*\sigma_{*,1})\sigma_{*,2}} > \frac{(\bar{b}_*(k))^2 \sigma_{*,2}^2}{\bar{b}_*(k)(\sigma_{*,3} + Kc\sigma_{*,2}\sigma_{*,1}) + Kc\sigma_{*,2}}, \quad (74)$$

The function

$$f(x) = \frac{x^2}{x + a} \quad (75)$$

has

$$f'(x) = \frac{x^2 + 2xa}{(x+a)^2}, \; f''(x) = \frac{(2x + 2a)(x+a) - 2(x^2 + 2xa)}{(x+a)^3} = \frac{2a^2}{(x+a)^3} > 0$$

and hence it is convex for $x > 0$. Thus,

$$\frac{1}{K} \sum_k \frac{b_*(k)^2 \sigma_{*,2}^2}{b_*(k)\sigma_{*,3} + c(1 + \bar{b}_*\sigma_{*,1})\sigma_{*,2}} > \frac{K^{-2}\bar{b}_*^2 \sigma_{*,2}^2}{K^{-1}\bar{b}_*\sigma_{*,3} + c(1 + \bar{b}_*\sigma_{*,1})\sigma_{*,2}}$$

which is equivalent to

$$\sum_k \frac{b_*(k)^2 \sigma_{*,2}^2}{b_*(k)\sigma_{*,3} + c(1 + \bar{b}_*\sigma_{*,1})\sigma_{*,2}} > \frac{\bar{b}_*^2 \sigma_{*,2}^2}{\bar{b}_*\sigma_{*,3} + Kc(1 + \bar{b}_*\sigma_{*,1})\sigma_{*,2}}$$

and the claim follows. □

## B  Benchmark Algorithms

In the simulation experiments of Section 4 we compare DRE to two benchmark algorithms. We implement the first, DNN, by selecting the architecture with best validation performance among 16 candidate networks. One of these networks aims at harvesting the strength of double descent. It uses a DNN with 7 layers and gradually halves the number of neurons in each successive layer, beginning $2^9$ neurons in the first layer and concluding with $2^3$ neurons in the last layer. We train this network with ADAM, an initial learning rate of 0.01, and batch size equal to the training size. We train the network for up to $12 \times 10^4$ epochs. We train the remaining 15 networks for 200 epochs with early stopping based on the validation sample. The network architecture is defined as $J$ layers with $J \in \{2, ..., 6\}$, and neuron count halving from $2^{J+3}$ in the first layer to $2^3$ in the last. For each architecture, we consider three different batch sizes, 64, 32, and 16. In each data sample the network that maximizes the performance measure on the validation sample.

The second benchmark model that we implement is random forests regression. We train six different forests with number of trees $N \in \{10, 100, 500, 1000, 5000, 10000\}$. We train the model to minimize the MSE and expand the trees until the leaves are pure or contain less than two samples. We select the best random forest based on its performance on the validation sample.